%% file: main.tex
\documentclass[nohyperref]{article}

\usepackage{hyperref}

\usepackage[accepted]{icml2022}

\usepackage{amsmath}
\usepackage{amssymb}
\usepackage{mathtools}
\usepackage{amsthm}

\usepackage{enumitem}

\usepackage[capitalize,noabbrev]{cleveref}

\theoremstyle{plain}
\newtheorem{theorem}{Theorem}[section]
\newtheorem{proposition}[theorem]{Proposition}
\newtheorem*{proposition*}{Proposition}

\theoremstyle{definition}

\theoremstyle{remark}

\usepackage[textsize=tiny]{todonotes}

\input{math_commands.tex}

\usepackage{wrapfig}
\usepackage{tikz-cd}
\usetikzlibrary{cd}
\usepackage{url}            
\usepackage{booktabs}       
\usepackage{amsfonts}       
\usepackage{nicefrac}       
\usepackage{microtype}      
\usepackage{graphicx}
\usepackage{caption}
\usepackage{subcaption}
\usepackage{siunitx}
\usepackage{amssymb}
\usepackage{multirow}
\usepackage[export]{adjustbox}

\newcommand{\cZ}{\mathcal{Z}}
\newcommand{\cA}{\mathcal{A}}
\newcommand{\mvdot}{\cdot}

\newcommand{\DIE}{\mathrm{DIE}}

\definecolor{lightgrey}{rgb}{0.43,0.43,0.43}

\icmltitlerunning{Learning Symmetric Embeddings for Equivariant World Models}

\begin{document}

\twocolumn[
\icmltitle{Learning Symmetric Embeddings for Equivariant World Models}



\icmlsetsymbol{equal}{*}

\begin{icmlauthorlist}
\icmlauthor{Jung Yeon Park}{equal,neu}
\icmlauthor{Ondrej Biza}{equal,neu}
\icmlauthor{Linfeng Zhao}{neu}
\icmlauthor{Jan Willem van de Meent}{uva,neu}
\icmlauthor{Robin Walters}{neu}
\end{icmlauthorlist}

\icmlaffiliation{neu}{Khoury College of Computer Sciences, Northeastern University, Boston, MA, USA}
\icmlaffiliation{uva}{Informatics Institute, University of Amsterdam, Amsterdam, Netherlands}

\icmlcorrespondingauthor{Jung Yeon Park}{park.jungy@northeastern.edu}
\icmlcorrespondingauthor{Ondrej Biza}{biza.o@northeastern.edu}
\icmlcorrespondingauthor{Robin Walters}{r.walters@northeastern.edu}

\icmlkeywords{equivariant, symmetry, contrastive loss, world models, transition, representation theory, generalization}

\vskip 0.3in
]



\printAffiliationsAndNotice{\icmlEqualContribution} 

\begin{abstract}
Incorporating symmetries can lead to highly data-efficient and generalizable models by defining equivalence classes of data samples related by transformations. However, characterizing how transformations act on input data is often difficult, limiting the applicability of equivariant models. We propose learning symmetric embedding networks (SENs) that encode an input space (e.g. images), where we do not know the effect of transformations (e.g. rotations), to a feature space that transforms in a known manner under these operations. This network can be trained end-to-end with an equivariant task network to learn an explicitly symmetric representation. We validate this approach in the context of equivariant transition models with 3 distinct forms of symmetry. Our experiments demonstrate that SENs facilitate the application of equivariant networks to data with complex symmetry representations. Moreover, doing so can yield improvements in accuracy and generalization relative to both fully-equivariant and non-equivariant baselines.
\end{abstract}

\vspace{-1.00\baselineskip}
\section{Introduction}

Symmetry has proved to be a powerful inductive bias for improving generalization in supervised and unsupervised learning. A symmetry group defines equivalence classes of inputs and outputs in terms of transformations that can be performed on the input along with corresponding transformations for the output. 
Recent years have seen many proposed equivariant models that incorporate symmetries into deep neural networks \citep{cohen2016group, cohen2016steerable, cohen2019gauge,  weiler2019e2cnn, weiler2018learning, kondor2018generalization, bao2019equivariant, worrall2017harmonic}. This results in models that are often more parameter efficient, more sample efficient, and safer to use by behaving more consistently in new environments. 

\begin{figure}[!t]
    \centering
\includegraphics[trim=0 0 0 0,clip,width=0.48\textwidth]{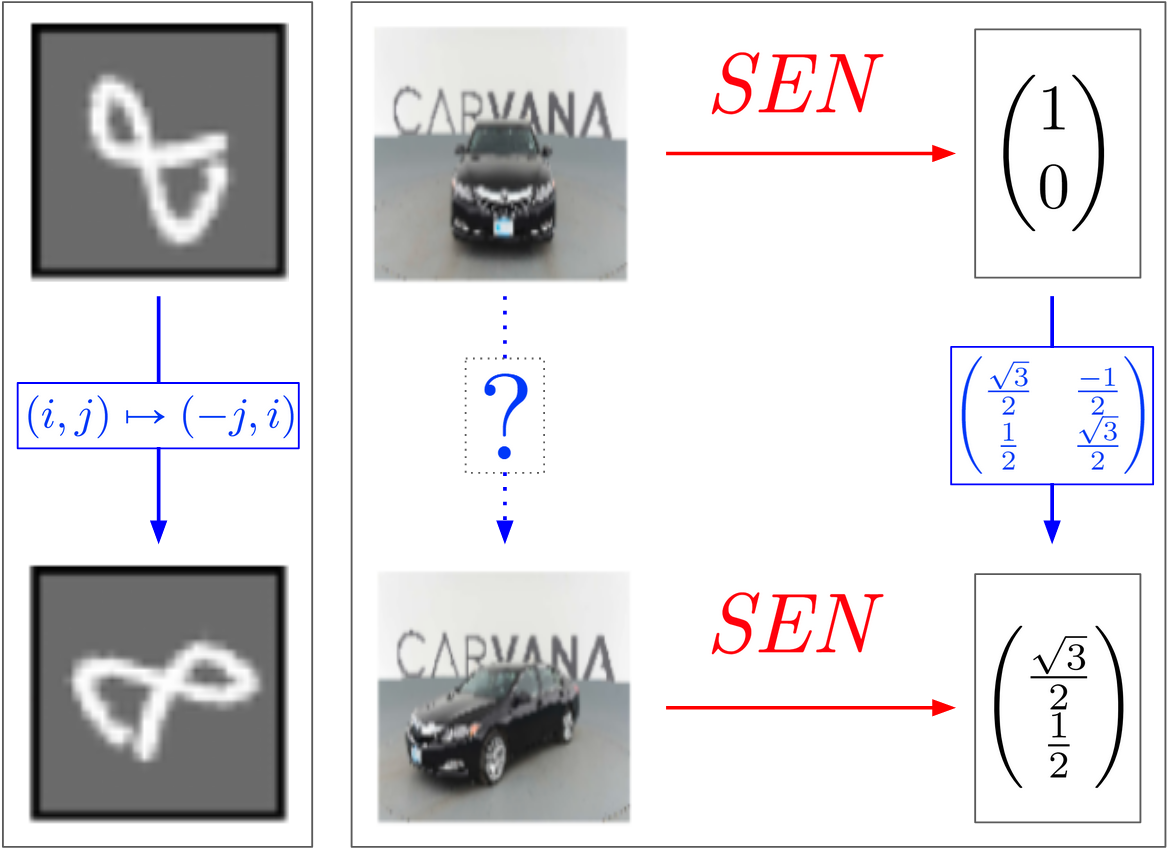}
    \caption{Equivariant networks consider transformations of inputs that are easy to compute, such as in-plane rotations for MNIST digits. This paper considers transformations that are difficult to compute, such as rotations of 3D objects like cars \citep{carvana}. Symmetric embedding networks learn representations which are transformed simply.}
    \label{fig:transform}
\end{figure}

However, the applicability of equivariant models is impeded in that it is not always obvious how a symmetry group acts on input data. For example, consider the pairs of images in Figure~\ref{fig:transform}. On the left, we have MNIST digits where a 2D rotation in pixel space induces a corresponding rotation in feature space. Here an $E(2)$-equivariant network achieves state of the art accuracy \citep{weiler2019e2cnn}. In contrast, exploiting the underlying symmetry is challenging for the images on the right, which are of the same object in two orientations. While there is also an underlying symmetry group of rotations, it is not easy to characterize the transformation in pixel space associated with a particular rotation.


In this paper, we consider the task of learning symmetric representations of data in domains where transformations cannot be hard-coded, i.e.~the \emph{group action} is unknown. We train a network that maps an input space, for which the group action is difficult to characterize, onto a latent space, where the action is known. We refer to this network as a symmetric embedding network (SEN). Our goal is to learn an SEN that is equivariant: for any pair of inputs related by a transformation in input space, the outputs should be related by a corresponding transformation in feature space.



Learning the group action from data requires supervision or inductive biases. In certain domains we can learn an SEN in a supervised manner by pairing it with an equivariant classifier. We demonstrate the feasibility of this approach in Section~\ref{sec:labeling}. However, our main interest is learning SENs in domains where direct supervision is not available. As a concrete instantiation of this setting, we focus on world models, i.e.~models that encode the effects of actions in the state space of a Markov decision process (MDP). We propose a meta-architecture that pairs an SEN with an equivariant transition network, which are trained jointly by minimizing a contrastive loss. The intuition in this approach is that the symmetry group of the transition model can help induce an approximately equivariant embedding network.

The idea of training an SEN in the form of a standard network with an equivariant task network has, to our knowledge, not previously been proposed or demonstrated. To test this idea, we consider 5 domains with 3 different symmetry groups, and 3 different equivariant architectures (see \autoref{tab:exp_summary} for more details). While not the main contribution of this paper, these domains do require innovations in architecture design. Most notably, we combine message passing neural networks (MPNNs) with $C_4$-convolutions in domains with multiple objects, resulting in a novel architecture which has been concurrently proposed in \cite{brandstetter2022geometric}. However, our primary contribution is to demonstrate that SENs can extend the applicability of equivariant networks to new domains with unknown group actions.  


We summarize our contributions as follows:
\vspace{-0.5em}
\begin{itemize}[leftmargin=*,itemsep=2pt,parsep=2pt,topsep=4pt]
    \item We propose SENs that map from input data, for which symmetries are difficult to characterize, to a feature space with a known symmetry. This network implicitly learns the group action in the input space.
    \item We show proof-of-concept results for supervised learning of SENs on sequence labeling.
    \item Using world models as a test case, we show SENs can be trained end-to-end by minimizing a contrastive loss. We develop 5 domains with 3 different symmetry groups using architectures that are representative of (and improve upon) the state of the art in equivariant deep learning.
    \item Our experiments show that world models with SENs can make equivariant architectures applicable to previously inaccessible domains and can yield improvements in accuracy and generalization performance.
\end{itemize}

\section{Related Work}

\paragraph{Equivariant Neural Networks} A multitude of equivariant neural networks have been devised to impose symmetry with respect to various groups across a variety of data types. These require that the group $G$ is known and the the group action on input, output, and hidden spaces is explicitly constructed.
Examples include $G$-convolution \citep{cohen2016group}, $G$-steerable convolution \citep{cohen2016steerable, weiler2019e2cnn}, tensor product and Clebsch-Gordon decomposition \citep{thomas2018tensor}, or convolution in the Fourier domain \citep{esteves2018polar}.  These models have been applied to gridded data \citep{weiler2019e2cnn}, spherical data \citep{cohen2018spherical}, point clouds \citep{dym2020universality}, 
and sets \citep{maron2020learning}.
%
They have found applications in many domains including molecular dynamics \citep{anderson2019cormorant}, particle physics \citep{bogatskiy2020lorentz},
and trajectory prediction \citep{walters2020trajectory}. In particular, \citet{ravindran2004algebraic} consider symmetry in the context of Markov Decision Processes (MDPs) and \citet{van2020mdp} construct equivariant policy networks for policy learning. Our work also considers MDPs with symmetry but focuses on learning equivariant world models (see Appendix \ref{app:mdp_symmetry}).


\paragraph{Learning Symmetry} 
Our work occupies a middle ground between equivariant neural networks with known group actions and symmetry discovery models. Symmetry discovery methods attempt to learn both the group and actions from data. For example, \citet{zhou2020meta} learn equivariance by learning a parameter sharing scheme using meta-learning. \citet{dehmamy2021automatic} similarly learn a basis for a Lie algebra generating a symmetry group while simultaneously learning parameters for the equivariant convolution over this symmetry group.  \citet{benton2020learning} propose an adaptive data augmentation scheme, where they learn which group of spatial transformations best supports data augmentation.

\citet{higgins2018definition} define disentangled representations based on symmetry, with latent factors considered disentangled if they are independently transformed by commuting subgroups.
Within this definition, \citet{quessard} learn the underlying symmetry group by interacting with the environment, where the action space is a group of symmetry transformations. Except for the 3D Teapot domain, 
we handle the more general case where the action space may be different from the symmetry group. Their latent transition is given by multiplication with a group element, whereas our transition model may be an equivariant neural network. 

\paragraph{Structured Latent World Models}
World models learn state representations by ignoring unnecessary information unrelated to predicting environment dynamics. Such models are frequently used for high-dimensional image inputs, and usually employ (1) reconstruction loss \citep{ha2018world,NIPS2015_a1afc58c,hafner2019learning,hafner2020mastering} or (2) constrastive loss. Minimizing the contrastive loss is known to be less computationally costly and can produce good representations from high-dimensional inputs \citep{oord2018representation,anand2019unsupervised,chen2020simple,laskin2020curl,van2020plannable}, thus we use it here. We take inspiration from \citet{kipf20}, who learn object-factored representations for structured world modeling with GNNs, which respect $S_n$ permutation symmetry. \citet{velickovic21reasoning} used a similar approach in Reasoning-Modulated Representations (RMR). RMR pre-train a transition model (“a processor”) from abstract data that compactly describes the underlying dynamics of the modelled system. In contrast, with the exception of 3D Teapot, SEN does not assume access to the underlying state or dynamics of the environment. We assume only that the group representation is known in the latent space, which cannot fully specify the system dynamics, and use this information as an inductive bias.
%

\section{Illustrative Example: Sequence Labeling}
\label{sec:labeling}

Our goal is to use an equivariant task network as an inductive bias for learning an SEN. While the SEN is itself not equivariant by construction, we may be able to learn an equivariant SEN by training both networks end-to-end. To validate 
this approach, we first
consider
a simple supervised learning problem in the form of a sequence labeling task.


In this simulated task, we detect local maxima in time series. Our training data (Figure~\ref{fig:sine1D_data}) comprises $N=10000$ sine waves with $T=100$ points, where each time series $x_n$ is shifted using a random offset $u_n$,
\begin{align}
    x_{n,t} &= \sin \left(\frac{4 \pi t}{100} \!+\! u_n \right), &
    u_n &\sim \text{Unif}\left(\left[-\frac{\pi}{2},  \frac{\pi}{2}\right] \right).
\end{align}
For each time point $x_{n,t}$, there is a label $y_{n,t} \in \{0,1\}$ indicating whether the point is a local maximum.


This domain has clear translational equivariance with known action: shifting inputs $x_{n}$ by $k$ time points result in shifting predictions $y_{n}$ by $k$ time points as well. For this reason, 1D convolutions are commonly used in sequence labeling \cite{pmlr-v32-santos14, ma-hovy-2016-end}. 

To test whether we can learn the group action, we compose a fully-connected (FC) layer, which acts as our non-equivariant SEN, with two translation-equivariant 1D convolutional layers. We compare this network against a non-equivariant network with three FC layers. Both networks use ReLU activations and one kernel for both convolutional layers for more interpretable visualizations. 

\begin{figure}[!t]
    \centering
    \begin{subfigure}[t]{0.32\columnwidth}
        \centering
        \includegraphics[width=\columnwidth]{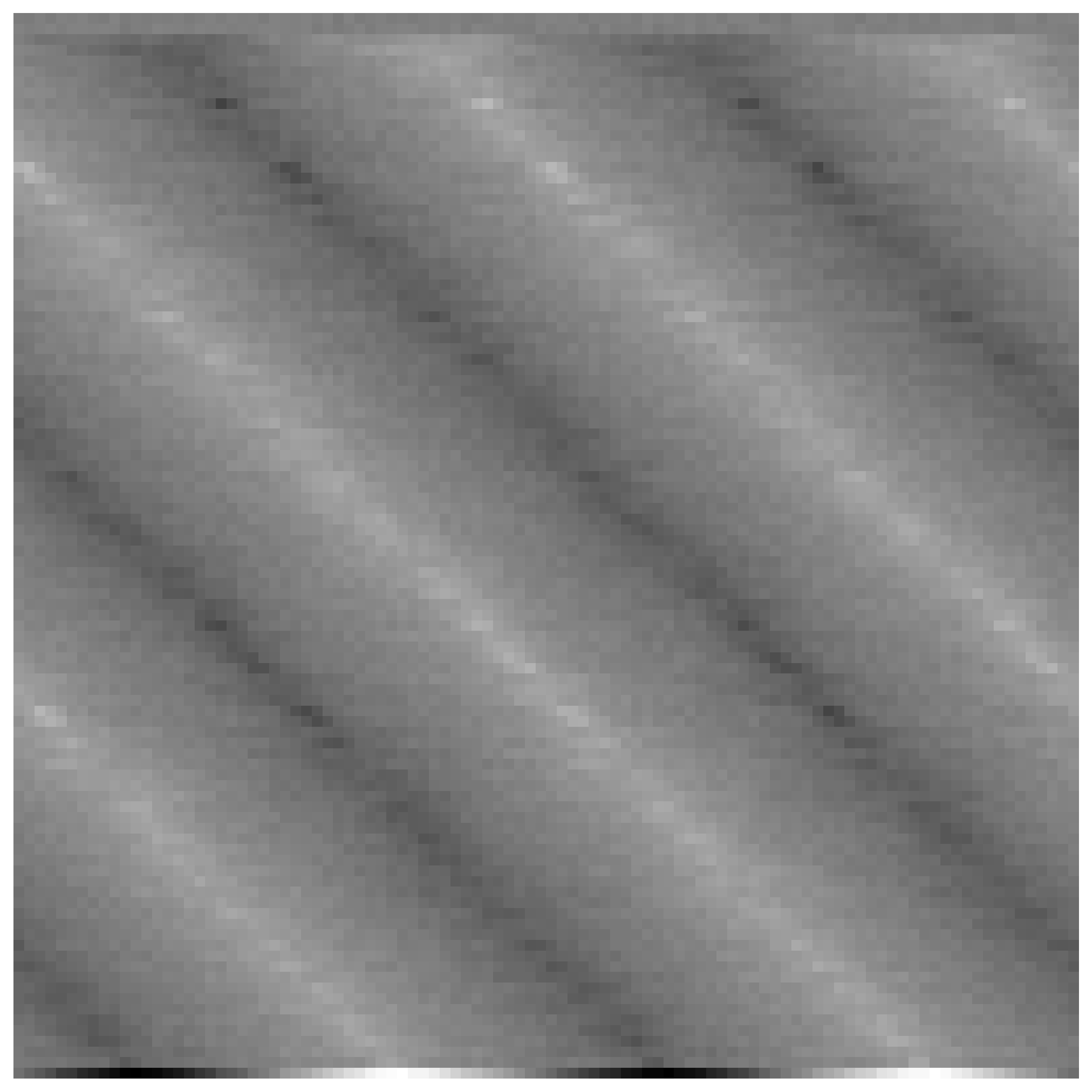}
        \caption{FC + Conv}
     \end{subfigure}
    \begin{subfigure}[t]{0.32\columnwidth}
        \centering
        \includegraphics[width=\columnwidth]{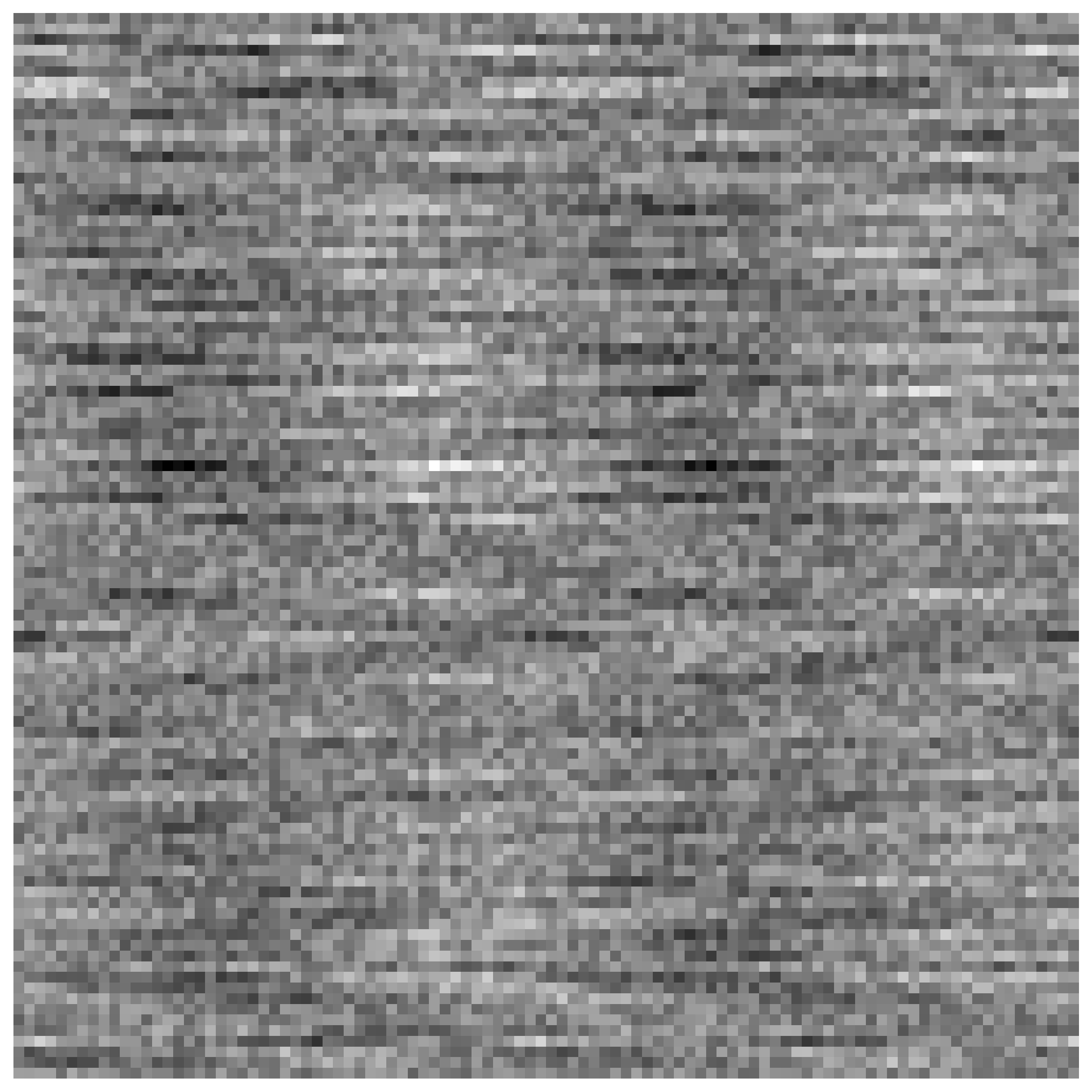}
        \caption{FC only}
     \end{subfigure}
      \begin{subfigure}[t]{0.32\columnwidth}
        \centering
            \includegraphics[width=\columnwidth]{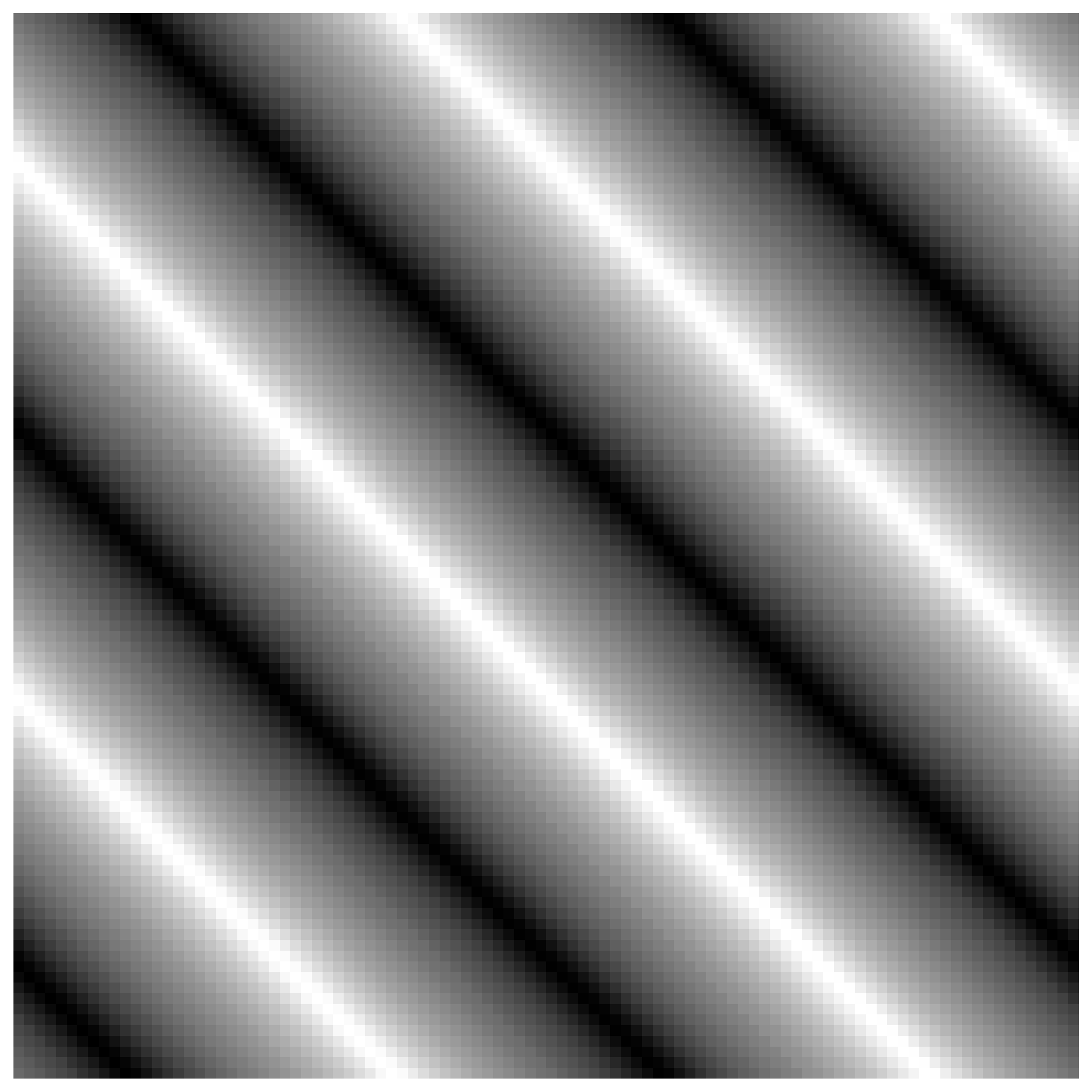}
        \caption{Equivariant FC}
     \end{subfigure}
    \caption{First FC layer weights for the (a) FC+Conv, and (b) FC only networks. (c) shows a perfect shift equivariant FC layer. The FC+Conv network learns shift equivariance.}
    \label{fig:sine1D_weights}
\end{figure}

Figure~\ref{fig:sine1D_weights} shows the weights for the first FC layer in both networks after end-to-end supervised training. We see that the learned weights in the FC+Conv model exhibit an approximate circulant structure (Fig.~\ref{fig:sine1D_weights}a), i.e. each column is shifted with respect to the preceding column. This is in excellent agreement with the idealized form that we would expect for a perfectly-equivariant layer (Fig.~\ref{fig:sine1D_weights}c). By contrast, the weights in the non-equivariant model (Fig.~\ref{fig:sine1D_weights}b) do not exhibit the same structure.

\section{Symmetric Embeddings for World Models}

The supervised learning results on toy data are encouraging: an equivariant task network can indeed induce an approximately equivariant SEN, which implicitly learns the group action in the input space. To demonstrate the feasibility of learning SENs in more challenging domains we consider world models. 
These models are an excellent use case for equivariant representation learning. Interactions with the physical world often exhibit symmetries, such as permutation equivariance (when interacting with multiple objects), or rotational and translational symmetries (when interacting with individual objects). Incorporating these symmetries can aid generalization across the combinatorial explosion of possible object arrangements, which grows exponentially with the number of objects in a scene. 

World models are also a good test bed for learning SENs in that they allow us to control the difficulty of the learning problem. In an equivariant world model there are three interrelated notions of ``action'': (1) the \emph{MDP} action in the world model, (2) the \emph{learned} action of the symmetry group on the input space, and (3) the \emph{known} action of the symmetry group in the latent space. In certain domains there is a direct correspondence between these notions, such as when MDP actions perform rotations on a single object. In other domains the correspondence will be more indirect, such as when MDP actions apply forces to joints in a robot arm. 
The MDP action in a world model hereby provides a form of ``distant'' supervision that either directly or indirectly relates to the underlying symmetry.

To establish notation, we first define the difference between an abstract symmetry group with known action and one with unknown action. We then define a meta-architecture for contrastive training of SENs and equivariant world models, and discuss implementations of this architecture for specific domains with different underlying symmetries.


\begin{figure*}[!t]
    \centering
    \includegraphics[width=0.9\textwidth]{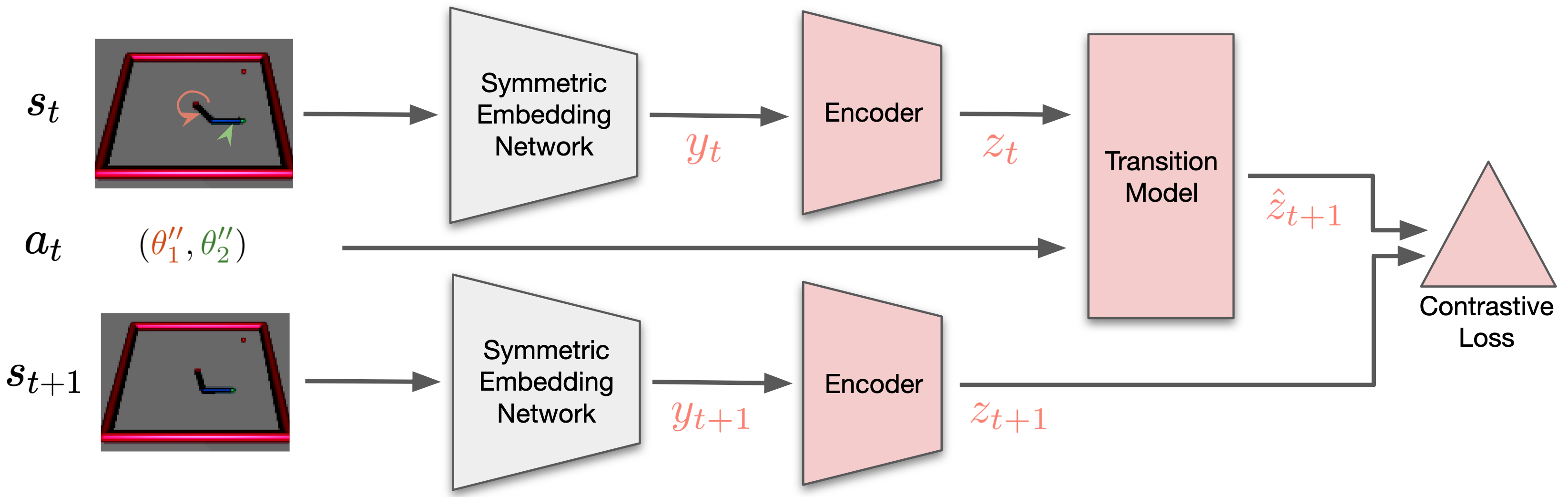}
    \caption{Diagram of the model architecture of our $G$-equivariant world model on the \emph{Reacher} domain with $G=D_4$ symmetry. The features in light red have an explicit $G$-action $\rho$.  The networks in light red are $G$-equivariant.
    The MDP actions have $G$-representation type $\rho_{\mathrm{flip}}$ meaning they are reversed in sign by reflections and unaltered by rotations. The Symmetric Embedding Network is a CNN and the Encoder and Transition model are $E(2)$-CNNs with fiber group $D_4$.}
    \label{fig:gwm}
\end{figure*}

\subsection{Preliminaries} 

\paragraph{Group Actions} 
A group is a set with a binary operation that satisfies associativity, $g_1 \cdot (g_2 \cdot g_3) = (g_1 \cdot g_2) \cdot g_3$, existence of an identity, $g \cdot 1 = 1 \cdot g = g$, and existence of an inverse, $g^{-1} \cdot g = g \cdot g^{-1} = 1$. 
An \emph{action} of a symmetry group associates a transformation with each $g \in G$. We define an action on a set $S$ as a map $a \colon G \times S \to S$ that is compatible with composition of group elements, which is to say that $a(g_1 \cdot g_2, s) = a(g_1, a(g_2,s))$.

If $S$ is a vector space $\mathbb{R}^n$ and the map $a(g,\cdot)$ on $S$ is linear, then we say that $a$ is a \emph{group representation}. This representation associates an $n\!\times\!n$ matrix with each $g \in G$, which we denote $\rho(g) = a(g,\cdot)$. The same group may have different actions on different sets.  For example, the cyclic group $C_4 = \lbrace 1,g,g^2,g^3 \rbrace$ has a simple representation $\rho_{\mathrm{std}}$ by $2\!\times\!2$-rotation matrices on vectors in $\mathbb{R}^2$ but a more complicated action by $28^2\!\times\!28^2$ matrices on images in MNIST. 


\paragraph{Equivariant Networks and Equivariance Learning}
Given a group $G$ with representations $\rho_X$ and $\rho_Y$ acting on $X$ and $Y$, we say that a function $f \colon X \to Y$ is \emph{equivariant} if, for all $x \in X, g\in G$,
\begin{equation}
    f\big(\rho_X(g) \mvdot x\big) = \rho_Y(g) \mvdot f(x).
\end{equation}
This means that group transformations commute with function application; transforming the input before application of $f$ is equivalent to transforming the output after application. The mapping $f$ thus preserves the symmetry group $G$ but alters the way in which the group acts.     
%

Equivariant neural networks define parametric families of equivariant functions $f$ by composing layers that are individually equivariant with respect to the same group. To ensure that equivariance can be satisfied by construction for any choice of network weights, these networks require that we explicitly know both $\rho_X$ and $\rho_Y$. An example would be classification of rotated MNIST digits, as in Figure~\ref{fig:transform}. 

In this paper, we are interested in cases where we have a known output action $\rho_Y$, but the input action $\rho_X$ is not known, as with the images of rotated cars in Figure~\ref{fig:transform}. In this setting, we are interested in learning equivariance using an unconstrained network $\hat{f}$, which we refer to as a symmetric embedding network (SEN). Given a triple $x_1,x_2,g$ such that $x_2 = \rho_X(g) x_1$, this network should satisfy
\begin{align}
    \hat{f}(x_2) &\approx \rho_Y(g) \cdot \hat{f}(x_1).
\end{align}
In other words, our goal is to learn a network $\hat{f}$ that is not equivariant by construction, but is as equivariant as possible. 

\paragraph{Equivariant World Models} World models define a transition function $T: \gS \times \gA \to \gS$ on a state space $\gS$ and action space $\gA$, which outputs the next state $s' = T(s,a)$ given the current state $s \in \gS$ and action $a \in \gA$. In an equivariant world model, we assume a symmetry group $G$ which jointly transforms states and actions by representations $\rho_\gS$ and $\rho_\gA$ respectively.  For example, in the 2D shapes domain shown in Table \ref{tab:exp_summary}, rotation by $\pi/2$ moves the blocks and permutes the actions $\rho_\gA(\mathrm{up})=\mathrm{left}$.  The transition function is equivariant with respect to $G$ in the sense that    
\begin{align}
    \label{eq:transition-equivariance}
    T\big(\rho_{\gS}(g) \!\cdot\! s, \rho_{\gA}(g) \!\cdot\! a \big) = \rho_{\gS}(g) \cdot T(s, a).
\end{align}
As with other equivariant approaches, recent work on equivariant world models has required that both $\rho_\gS$ and $\rho_\gA$ are known \cite{van2020mdp}.

\subsection{Meta-Architecture and Contrastive Loss}

In this paper, we use SENs to define approximately-equivariant world models that can be trained without access to $\rho_{\gS}$. To do so, we define a meta-architecture that combines a symmetric embedding network with an equivariant world model, which is illustrated in Figure~\ref{fig:gwm}. This architecture comprises three domain-dependent components:

1. A symmetric embedding network $S \colon \gS \to \gY$, which maps states in a pixel-space $\gS$ to an intermediate space $\gY$ for which an explicit symmetry group action $\rho_{\gY}$ is known. 

2. An equivariant encoder $E \colon \gY \to \gZ$, which extracts the subset of features that are necessary to predict dynamics in a latent space $\gZ$ with a known group action $\rho_{\gZ}$.

3. An equivariant transition model $T: \gZ \times \gA \to \gZ$ which serves as an inductive bias by defining dynamics that satisfy the relation in Equation~\ref{eq:transition-equivariance} with respect to the known group representations $\rho_{\gZ}$ and $\rho_{\gA}$.

 

We employ the self-supervised contrastive loss introduced by \citet{kipf20} for training. We assume access to a dataset collected offline of triplets $(s,a,s')$ consisting of the current state $s$, the action $a$, and the next state $s'$. We combine this ground truth transition triplet with a negative state $s''$, which is randomly sampled from triplets within the minibatch.
The contrastive loss is
\begin{align*}
    \mathcal{L} \:=~ 
     \| T(z,a) - z' \| 
     + \alpha \: \mathrm{max}\big(\beta-\| T(z,a) - z'' \|,0\big),
\end{align*}
where $z= E(S(s))$, $z'=E(S(s'))$, and $z''=E(S(s''))$. Minimizing this loss pushes $T(z,a)$ towards $z'$ and away from the negative sample $z''$.  

\subsection{Environments and Architectures}


\begin{table*}[t] 
\def\arraystretch{2.3}%
\centering
\resizebox{0.9\textwidth}{!}{
\begin{tabular}{l cccc}
& \includegraphics[width=0.18\textwidth]{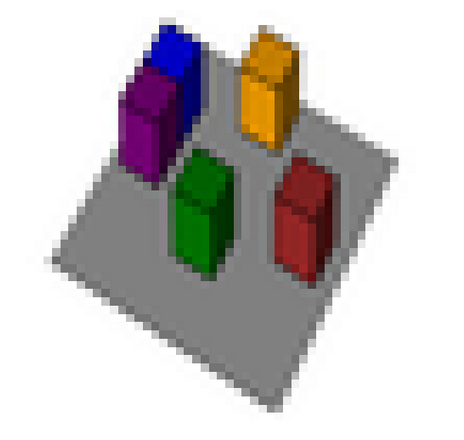} & \includegraphics[width=0.17\textwidth]{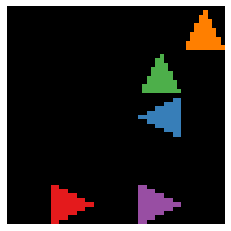} & \includegraphics[width=0.2\textwidth]{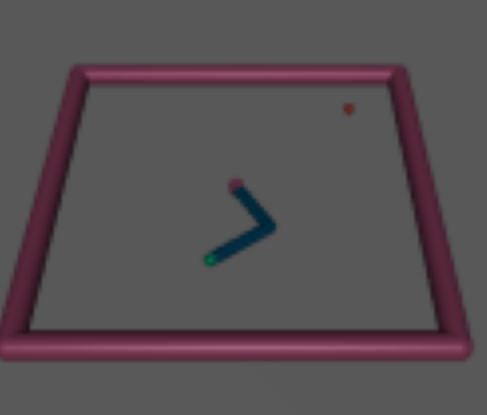} & \includegraphics[width=0.17\textwidth]{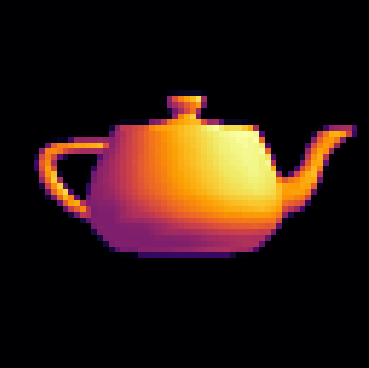} \\
\toprule
Environment & 2D Shapes \& 3D Blocks   & Rush Hour & Reacher & 3D Teapot \\\midrule
Observation $s$ & 50x50x3 & 50x50x3 & 128x128x3x2 & 64x64x1 \\
Action $a$ &   $\{$ up,right,down,left $\}$ & 
\def\arraystretch{1} \begin{tabular}{@{}c@{}} $\{$ fwd,left,back,right $\}$\end{tabular} & \def\arraystretch{1} \begin{tabular}{@{}c@{}} $(\phi_1'', \phi_2'') \in \mathbb{R}^2$ \\ (joint forces) \end{tabular} & $\mathrm{SO}(3)$ \\
Symmetry $G$  & \def\arraystretch{1}  \begin{tabular}{@{}c@{}}$C_4 \times S_5$ \\ ($\frac{\pi}{2}$ rot.; obj. perm.) \end{tabular} & \def\arraystretch{1}  \begin{tabular}{@{}c@{}}$C_4 \times S_5$ \\ ($\frac{\pi}{2}$ rot.; obj. perm.) \end{tabular} & \def\arraystretch{1}  \begin{tabular}{@{}c@{}} $D_4 \ltimes (\mathbb{R}^2,+)$ \\ ($\frac{\pi}{2}$ rot; flip; trans.) \end{tabular} & $\mathrm{SO}(3)$ \\
$\gZ$-rep: $\rho_{\gZ}$ & $(\rho_{\mathrm{std}},\mathbb{R}^2) \boxtimes (\rho_{\mathrm{std}}, \mathbb{R}^5)$ & \def\arraystretch{1}  \begin{tabular}{@{}c@{}} $(\rho_{\mathrm{std}} \oplus \rho_{\mathrm{reg}},\mathbb{R}^6)$ \\ $\boxtimes (\rho_{\mathrm{std}}, \mathbb{R}^5)$ \end{tabular} & $(\rho_{\mathrm{reg}},\mathbb{R}^8)^4 \boxtimes \rho_{\mathrm{triv}}$ & $gz$  (matrix mult.) \\
$\gA$-rep: $\rho_{\gA}$ & $(\rho_{\mathrm{reg}},\mathbb{R}^4) \boxtimes (\rho_{\mathrm{std}}, \mathbb{R}^5)$ & $(\rho_{\mathrm{triv}},\mathbb{R})^4$ & $(\rho_{\mathrm{flip}},\mathbb{R})^2$ & $gag^{-1}$\ (conjugation)  \\
SEN $S$ &  \def\arraystretch{1}  \begin{tabular}{@{}c@{}} 2-layer CNN (2D) \\ 4-layer CNN (3D) \end{tabular} & 2-layer CNN & 7-layer CNN  &  4 conv, 3 FC layers \\
Equ.\! Encoder $E$ &  \def\arraystretch{1}  MLP + $C_4$-conv & MLP + $C_4$-conv & \def\arraystretch{1} \begin{tabular}{@{}c@{}} 3 $E(2)$-conv, \\ 3-layer $D_4$-EMLP \end{tabular} & Id. (None) \\
Equ.\! Transition $T$ &  \def\arraystretch{1}  \begin{tabular}{@{}c@{}} \\ MPNN + $C_4$-conv \\ \cite{cohen2016group} \\  \cite{scarselli2008graph} \end{tabular} & MPNN + $C_4$-conv &  \def\arraystretch{1} \begin{tabular}{@{}c@{}} \\  EMLP, $E(2)$-CNN \\ \cite{weiler2019e2cnn}  \end{tabular}  &  \def\arraystretch{1} \begin{tabular}{@{}c@{}} Matrix \\ Multiplication \end{tabular} \\
\bottomrule
\end{tabular}
}
\caption{The symmetry and implementation for each domain.  See Appendix \ref{app:reps} for the $\rho$ definitions. 
\label{tab:exp_summary}
}  
\end{table*}
We consider 5 environments with varying symmetries. Table~\ref{tab:exp_summary} shows an overview of symmetries, representation types, and model architectures. The first two environments, 2D Shapes and 3D Blocks, are grid-worlds with 5 moving objects \citep{kipf20}. Rush Hour is a variant of 2D Shapes where objects move relative to their orientation. In these 3 domains, we consider symmetry to $\pi/2$ rotations ($C_4$) and object permutations ($S_5$). The fourth domain is a continuous control domain, the \texttt{Reacher-v2} MuJoCo environment \cite{todorov2012mujoco}, which is symmetric under rotations, flips ($D_4$), and translations. The last domain is a 3D teapot, where actions are 3D rotations in the group $\mathrm{SO}(3)$.  All environments use images as observed states. 

We here provide a high-level description of the transition model $T$, the encoder $E$, and the SEN $S$ for each environment. Additional details are in Appendices \ref{app:environ} and \ref{app:model}. 

\paragraph{Transition Model ($T$)} The transition model $T$ defines the main inductive bias. In 2D shapes, 3D blocks, and Rush Hour, we use a message-passing neural network. This defines an object-factored representation that is equivariant to permutations and models pairwise interactions (i.e.~movement of one object can be blocked by another object). We extend the architecture proposed by \citet{kipf20} to incorporate rotational symmetry using $C_4$ convolutions, resulting in a network similar to the one that has concurrently been proposed by \citet{brandstetter2022geometric}. 

The Reacher and 3D Teapot environments do not use model components that consider permutations. In Reacher, $T$ is an Equivariant MLP (EMLP) made with 1x1-convolutions in the $E(2)$-CNN framework \citep{weiler2019e2cnn}. 

\begin{figure}[!b]
    \centering
    \includegraphics[width=0.45\textwidth,trim=35 150 50 70,clip]{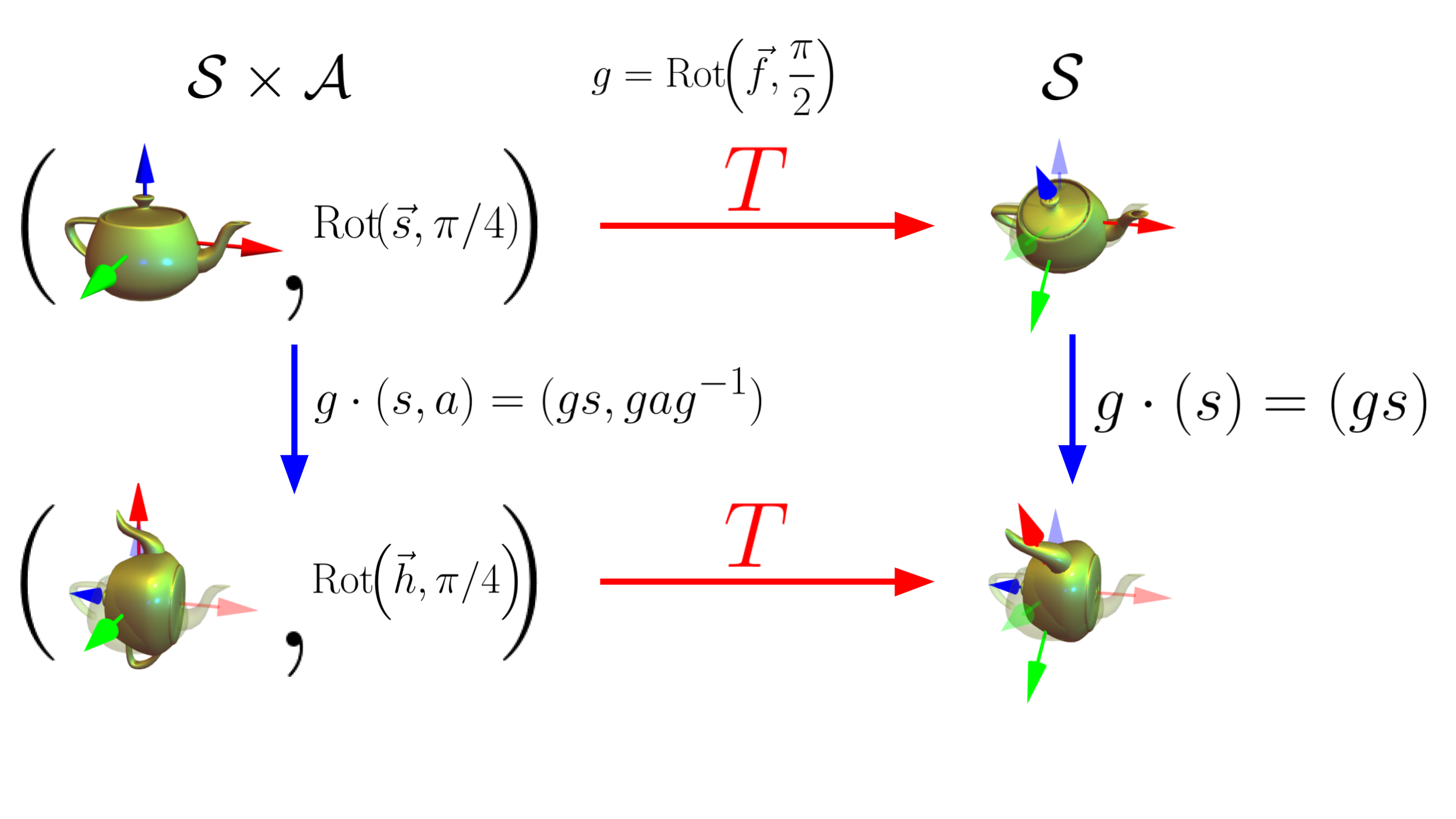}
    \caption{$\mathrm{SO}(3)$-equivariance for 3D Teapot. }
    \label{fig:teapot}
    \vspace*{-3mm}
\end{figure}

For 3D Teapot, the action space $\cA$, symmetry group $G$, and latent space $\cZ$ are all $\mathrm{SO}(3)$.
Since $\cZ$ is not a vector space, $\rho_\gA$ is a non-linear group action.
Semantically, the MDP action $a \in \mathrm{SO}(3)$ is a rotation matrix and the latent state $z \in \mathrm{SO}(3)$ is a positively-oriented orthogonal coordinate frame.  Though $\gZ = \gA = \mathrm{SO}(3)$, these interpretations lead to differing $G = \mathrm{SO}(3)$ actions with $\rho_\gZ(g)(z) = gz$ but $\rho_\gZ(g)(a) = gag^{-1}$ (see Figure~\ref{fig:teapot} for an illustration).  If $z$ is correctly learned, then $T$ can be implemented as a matrix multiplication $T_\gZ(z,a) = az$ which is equivariant, 
\begin{align}
    \begin{split}
    & T_\gZ(\rho_\gZ(g)(z),\rho_\gA(g)(a)) = \\
    & \qquad (gag^{-1})(gz) = gaz = \rho_\gZ(g)T_\gZ(z,a).
    \end{split}
\end{align}
This method, which we label \texttt{MatMul}, is similar to the one in \citet{quessard}, except in our framework the ground truth $a \in \mathrm{SO}(3)$ is provided to aid learning $z$. 

\paragraph{Equivariant encoder ($E$)} The encoder in object-centric environments is shared over all 5 objects and uses group convolution over $C_4$ \citep{cohen2016group}, thus achieving $C_4 \times S_5$-equivariance. In the Reacher environment, we combine 3 $E(2)$-equivariant layers with a 3-layer $D_4$-EMLP. For 3D Teapot, no encoder is needed. 





\paragraph{Symmetric Embedding Network ($S$)}
 The SEN in each environment is based on a convolutional network, an architecture well-adapted to our image inputs.  
 It maps the image $s$ to an image $y$ in which the $G$-action is easy to describe in terms of pixel manipulation. As each domain has different input size and downstream architectures, the SEN architecture also varies.

In the object-centric environments, the output $y$ is a down-sampled image with 5 channels. The action $\rho_\gY$ of $S_5$ permutes the channels while $C_4$ rotates the image. In the Reacher environment, $y$ is a down-sampled image which is rotated, flipped, and translated by $D_4 \ltimes (\mathbb{R}^2,+)$ via $\rho_\gY$.

In the case of the 3D teapot environment, we expect the SEN to detect the pose $z$ of the object in 3D.  We use a two-part network that directly encodes $y = z$ using a down-sampling CNN whose output is passed to an MLP, and converted to an element of $\mathrm{SO}(3)$. To force the output of the symmetric embedding network $y$ to be an element of $\mathrm{SO}(3)$, we have the last layer output 2 vectors $u,v \in \mathbb{R}^3$ and perform Gram-Schmidt orthogonalization to construct a positively oriented orthonormal frame (see Appendix \ref{app:model}). 
This method is also used by \citet{falorsi2018explorations}, who 
conclude 
it produces less topological distortion than alternatives such as quaternions.

\subsection{Generalizing over the MDP Action Space}

In settings where data collection is costly, equivariance can improve sample efficiency and generalization. 
While it is difficult to generalize over high-dimensional states without explicit symmetry, the MDP actions are low dimensional and have clear symmetry. Furthermore, the MDP action bypasses the non-equivariant $S$ and is passed directly to $T$, which is explicitly equivariant. This means we can train using only a proper subset $\gA' \subset \gA$ of the action space and then test on the entire $\gA$. In other words, our model has the added benefit of generalizing to unseen actions when trained on only a fraction of data, which we demonstrate in Section~\ref{exp:generalization}. We state a proposition that bounds the model error over the entire action space when the model is trained on the subset $\gA'$. (proof in Appendix \ref{app:prop-proof}).






\begin{proposition}\label{prop:gen-limited-actions}
Let $\gA' \subset \gA$.  Assume $\rho_\gA(G) \mvdot \gA' = \gA$, i.e. every MDP action is a $G$-transformed version of one in $\gA'$.  Consider $\gD'$ sampled from $\gS \times \gA' \times \gS$.  
Denote $\gD = G \cdot \gD' \subset \gS \times \gA \times \gS$ the set of all $G$-transforms of all of samples in $\gD'$ and $T_\gS(s,a) = T(E(S(s)),a)$.   
%
%
Assume a $G$-invariant norm and model error is bounded $\| T_\gS(s,a) - z' \| \leq \epsilon_1$ where $z' = E(S(s'))$ and 
equivariance errors are bounded $\| T_\gS(\rho_\gS(g)s,\rho_\gA(g)a) - \rho_\gZ(g) T_\gS(s,a)  \| \leq \epsilon_2$ and $\| E(S(\rho_\gS(g)s') - \rho_\gZ(g) z' \| \leq \epsilon_3$ for all $g \in G$ and all $(s,a,s') \in \gD'$.  Then model error over $\gA$ is also bounded $\| T_\gS(s,a) - z' \| \leq \epsilon_1 + \epsilon_2 + \epsilon_3$ for all $(s,a,s') \in \gD$. 
\end{proposition}





\section{Experiments}

For all experiments, we consider three types of models: (a) a non-equivariant model with no enforced symmetry, (b) a fully-equivariant model with 
$\rho_\gS$ chosen to be the closest explicit pixel-level transformation to the actual symmetry, and (c) our method.
For 3D Teapot, we forgo the fully equivariant baseline as it is hard to define a $\rho_\gS$ acting on the 2D image space which approximates the true group action. We instead include a comparison to Homeomorphic VAE \citep{falorsi2018explorations} which is trained to on images of teapots without any actions. As its latent space is the same as our model, we can use the \texttt{MatMul} transition model in order to predict the effect of actions.
We keep the total number of parameters comparable across all models by reducing the hidden dimensions for the equivariant networks. Other details are provided in Appendix~\ref{app:model} and \ref{app:training}. The code for our implementation is available\footnote{\url{https://github.com/jypark0/sen}}.

\begin{table*}[!t]
\centering
\resizebox{0.9\textwidth}{!}{
\begin{tabular}{llrrrrrr}
\toprule
& Model & \begin{tabular}{@{}c@{}} TH@1 \\ (yaw, \%)\end{tabular} & \begin{tabular}{@{}c@{}} TH@1 \\ (pitch, \%)\end{tabular} & 
\begin{tabular}{@{}c@{}} TH@1 \\ (roll, \%)\end{tabular} & \begin{tabular}{@{}c@{}} HH@1 \\ (1 step, \%)\end{tabular} & EE($S$) & 
\begin{tabular}{@{}c@{}} DIE \\ (1 step, $\times 10^{-2}$)\end{tabular} \\ 
\midrule
\multirow{3}{*}{3D Teapot} 
 & Homeomorphic VAE & $6.7$ & $10.0$ & $3.3$ & $0.9$ & $2.41$ & $0.68$ \\
 & None & $6.7{\color{gray}\scriptstyle \pm 6.7}$ & $60{\color{gray}\scriptstyle \pm 40}$ & $86.7{\color{gray}\scriptstyle \pm 6.6}$ & $93.9{\color{gray}\scriptstyle \pm 2.2}$ & $2.38{\color{gray}\scriptstyle \pm 0.04}$ & $3.41{\color{gray}\scriptstyle \pm 0.16}$\\
 & Ours (\texttt{MatMul}) & $100{\color{gray}\scriptstyle \pm 0.0}$ & $100{\color{gray}\scriptstyle \pm 0.0}$ & $100{\color{gray}\scriptstyle \pm 0.0}$ & $100{\color{gray}\scriptstyle \pm 0.0}$ & $0.05{\color{gray}\scriptstyle \pm 0.0}$ & $0.45{\color{gray}\scriptstyle \pm 0.01}$ \\
\bottomrule
\end{tabular}
}
\caption{Model performance on 3D Teapot.}
\label{tab:teapot}

\end{table*}



\subsection{Metrics}

To evaluate the model without state reconstructions, we use two types of metrics. The first are accuracy metrics adapted from \citet{kipf20} and the second are equivariance metrics to measure the degree of equivariance. 



\begin{figure}[!b]
    \centering
     \begin{subfigure}[b]{0.14\columnwidth}
        \centering
        \includegraphics[width=\textwidth,trim=0 0 0 0,clip]{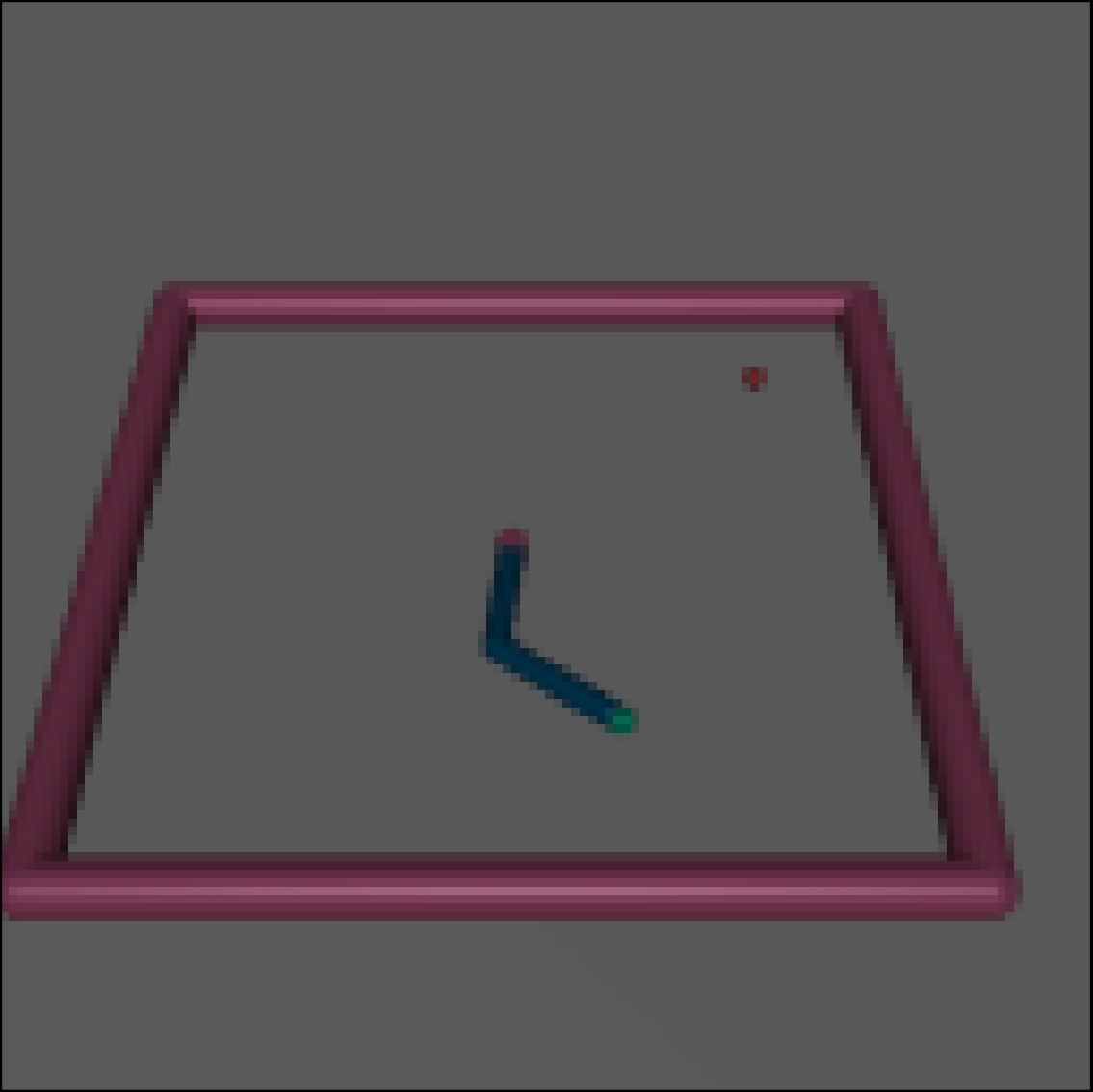}
    \end{subfigure}
    \hfill  
    \begin{subfigure}[b]{0.85\columnwidth}
        \centering
        \includegraphics[width=\textwidth,trim=0 0 0 0,clip]{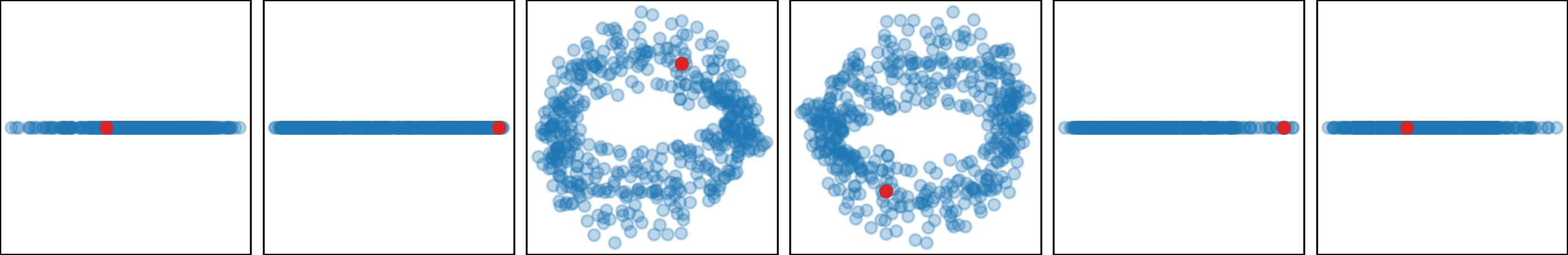}
     \end{subfigure}
    \caption{Reacher latent embeddings: a sample state $s$ (left) and its latent $z$ (right). The current $z$ is in red, blue points show other samples for reference. }
    \vspace{0.75\baselineskip}
    \label{fig:reacher_z_main}
    \hfill
    \centering
    \includegraphics[width=\columnwidth]{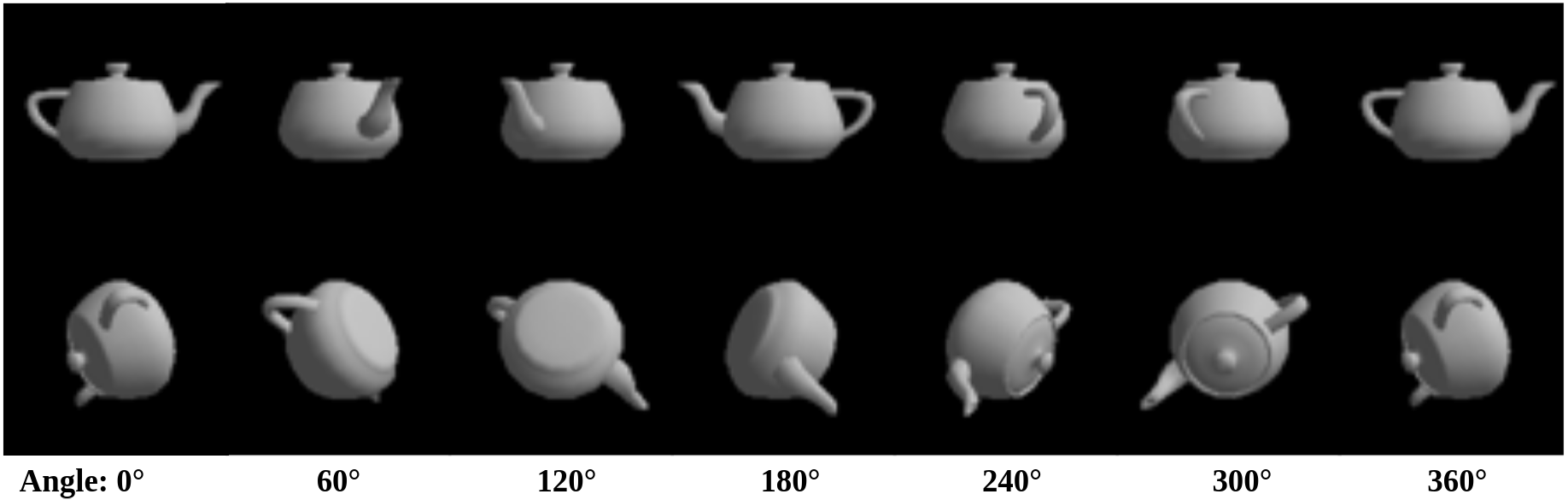}
    \caption{3D Teapot latent space traversal. True rotation (top) and our embeddings (bottom). The two sequences are offset by the arbitrary learned latent reference frame.}
    \label{fig:teapot_traversal}
\end{figure}

\paragraph{Hits, Hard Hits, Traversal Hits, and MRR} 
Given a dataset of triplets $(z,a,z')$, ranking metrics compute the $L_2$ distance between each predicted state $T(z,a)$ and all next states $z'$. The Hits at Rank $k$ (H@k) computes the proportion of triplets for which $T(z,a)$ is among the $k$-nearest neighbors of the corresponding next state $z'$. The mean reciprocal rank (MRR) is the average inverse rank. We also compute Hard Hits at Rank $k$ (HH@k), where we generate negative samples of states $s'_n$ close to the true next state $s'$ and compute the proportion of samples where the distance to $z'$ is lower than the distance to $z'_n$. This is a harder version of H@k, as the model must distinguish between similar negative samples and the true positive sample. For Traversal Hits (TH@k), which we use for Teapot experiments, we use $30$ $\frac{\pi}{15}$ increments along three axes of rotation (yaw, pitch, roll) to be the negative states $s'_n$. We measure whether $z'$ at each increment can be distinguished from the $z'_n$ of the other points along the traversal. For example, the traversal shown in Figure \ref{fig:teapot_traversal} reaches 100\%, whereas a model mapping every increment into the same latent state has TH@1 = 0\%.



\paragraph{Equivariance Error ($\mathrm{EE}$)} To evaluate the degree to which the learned SEN is equivariant, we generate triplets $(s, a, s')$ for which a known element $g$ acts on the state $s$. 
This yields images $s$ and $s' = \rho_{\gS}(g) \mvdot s$ during generation, which allows us to compute the equivariance error,
\[
    \mathrm{EE} = \mathbb{E}_{s,g} \left[ |\rho_\gY(g) \mvdot S(s) - S(\rho_\gS(g) \!\mvdot\! s) | \right].
\]
\paragraph{Distance Invariance Error ($\mathrm{DIE}$)} The equivariance error can be computed when the output space is spatial and we can manually perform group actions on the outputs. However it cannot be applied to the latent space $\gZ$ in the case of non-equivariant models, since the group action on the latent space $\rho_\gZ$ cannot be meaningfully defined.

We therefore propose a proxy for the equivariance error using invariant distances. For a pair of input states $s,s'$,
an equivariant model $f$ will have the same distances $\|f(s) - f(s')\|$ and $\|f(gs) - f(gs')\|$ assuming the action of $G$ is norm preserving, as it is for all transformations considered in the paper. 
(The action $\rho_\gS$ is assumed.)
Due to the linearity of the action, $\|f(gs) - f(gs')\| = \|gf(s) - gf(s')\| = \|g(f(s) - f(s'))\| = \|(f(s) - f(s'))\|$. 
The distance invariance error is computed as 
\[
\mathrm{DIE} = \mathbb{E}_{s, s',g} \left[ \left| \|f(s) - f(s')\| - \|f(gs) - f(gs')\| \right| \right].
\]

\subsection{Model performance comparison}

\begin{table*}[t]
\centering
\resizebox{0.8\textwidth}{!}{
\begin{tabular}{cclrrrr}
\toprule
& \begin{tabular}[c]{@{}l@{}}Limited \\
Actions  $\mathcal{A}^{\prime}$\end{tabular} & Model & \begin{tabular}{@{}c@{}} H@1 \\ (10 step, \%)\end{tabular} & \begin{tabular}{@{}c@{}}MRR \\ (10 step, \%)\end{tabular} & EE($S$) & \begin{tabular}{@{}c@{}} DIE \\ (10 step, $\times10^{-3}$)\end{tabular} \\
\midrule
\multicolumn{1}{c}{\multirow{2}{*}{2D Shapes}} & {\multirow{2}{*}{\{up\}}} & CNN & $2.8{\color{gray}\scriptstyle \pm 0.6}$ & $5.3{\color{gray}\scriptstyle \pm 0.4}$ & $0.00{\color{gray}\scriptstyle \pm 0.0}$ & $0.19{\color{gray}\scriptstyle \pm 0.0}$ \\
\multicolumn{1}{c}{} &  & Ours/Full & $100{\color{gray}\scriptstyle \pm 0.0}$ & $99.9{\color{gray}\scriptstyle \pm 0.0}$ & $0.00{\color{gray}\scriptstyle \pm 0.0}$ & $0.00{\color{gray}\scriptstyle \pm 0.0}$ \\
\midrule
\multirow{3}{*}{3D Blocks} & \multirow{3}{*}{{\{up,right,down\}}} & None & $52.3{\color{gray}\scriptstyle \pm 14.}$ & $61.8{\color{gray}\scriptstyle \pm 13.}$ & $0.98{\color{gray}\scriptstyle \pm 0.2}$ & $181{\color{gray}\scriptstyle \pm 79.}$ \\
&  & Full & $83.7{\color{gray}\scriptstyle \pm 36.}$ & $86.0{\color{gray}\scriptstyle \pm 31.}$ & $0.81{\color{gray}\scriptstyle \pm 0.5}$ & $15{\color{gray}\scriptstyle \pm 9.1}$ \\
&  & Ours & $99.9{\color{gray}\scriptstyle \pm 0.0}$ & $100{\color{gray}\scriptstyle \pm 0.0}$ & $0.96{\color{gray}\scriptstyle \pm 0.3}$ & $5{\color{gray}\scriptstyle \pm 4.7}$ \\
\bottomrule
\end{tabular}
}
\vspace{-0.5\baselineskip}
\caption{2D Shapes, 3D Blocks generalization results. The models were trained on a limited set of actions $\cA'$ and evaluated on all actions. For 2D shapes, due to the simplicity of the environment, a simple CNN turns out to be equivariant, so both the baseline CNN and Ours/Full have an equivariant symmetric embedding network.}
\label{tab:blocks_skew}
\end{table*}

\begin{table}[t]
\centering
\resizebox{0.49\textwidth}{!}{
\begin{tabular}{lrrrr}
\toprule
& \begin{tabular}{@{}c@{}} H@10 \\ (1 step, \%)\end{tabular} & \begin{tabular}{@{}c@{}}MRR \\ (1 step, \%)\end{tabular} & EE($S$) & \begin{tabular}{@{}c@{}} DIE (1 step) \\ ($\times 10^{-1}$)\end{tabular} \\ 
\midrule
None & $86.5{\color{gray}\scriptstyle \pm 3.0}$ & $50.6{\color{gray}\scriptstyle \pm 3.1}$ & $1.22{\color{gray}\scriptstyle \pm 0.1}$ & $6.95{\color{gray}\scriptstyle \pm 1.5}$ \\
Full & $89.4{\color{gray}\scriptstyle \pm 11.}$ & $61.8{\color{gray}\scriptstyle \pm 13.}$ & $1.18{\color{gray}\scriptstyle \pm 0.1}$ & $4.87{\color{gray}\scriptstyle \pm 1.4}$ \\
Ours & $90.8{\color{gray}\scriptstyle \pm 4.5}$ & $59.4{\color{gray}\scriptstyle \pm 4.6}$ & $1.28{\color{gray}\scriptstyle \pm 0.1}$ & $4.95{\color{gray}\scriptstyle \pm 0.6}$ \\
\bottomrule
\end{tabular}
}
\vspace{-0.5\baselineskip}
\caption{Reacher generalization results. The models were trained on data where the second joint is constrained to be positive and evaluated on unconstrained data.}
\label{tab:reacher_skew}
\vspace{-\baselineskip}
\end{table}

Tables~\ref{tab:teapot}, \ref{tab:blocks_rush_hour}, and \ref{tab:reacher} summarize performance of models and baselines, with additional results in Appendix~\ref{app:experiments}.
In general, the ranking metrics show that all models are accurate on 3D blocks, Rush Hour, and Reacher. 

Surprisingly, the fully equivariant model performs very well even when the group action on the input space $\rho_\gS$ is not accurate.  Due to the skewed perspective, we can see that the simple pixel-level transformation maps training data to out-of-distribution images which are never seen by the model. We hypothesize that equivariance does not hamper its performance on training data, but only constrains its extrapolation capabilities to out-of-distribution samples. 

In Table~\ref{tab:teapot}, we observe that both ``None'' and our model are accurate on hard hits (HH@1), but ``None'' performs poorly on traversal hits (TH@1). This baseline is only sensitive to pitch and roll, while completely ignoring the yaw of the teapot. The Homeomorphic VAE results indicate that the model makes only coarse distinctions between different orientations of the teapot.

For the equivariance metrics, we can see that SEN-based models outperform baselines on $\DIE$ for 3D Blocks and on $\mathrm{EE}(S)$ for Teapot. For the other environments, the equivariance metrics are relatively similar for all models.

\vspace{-0.25\baselineskip}
\subsection{Latent visualizations} 

We visualize the latent embeddings $z$ of our model to qualitatively analyze the learned representations. Figure~\ref{fig:reacher_z_main} shows a sample from the evaluation dataset in both pixel and latent space. We factor the encoded state $z$ and next state $z'$ into irreducible $D_4$-representations (irreps) before visualizing (see \cite{hall2003lie}). Some irreps are 1-dimensional and are plotted as a line. The 2-dimensional irreps show a clear circular pattern, matching the joint rotations of the environment.

We also transform the embedding with the group action $\rho_\gZ$ and visualize the corresponding pixel-level outputs.
Figure~\ref{fig:teapot_traversal} shows the traversal of rotations in pixel and latent space for 3D Teapot. The latent space can choose its own base coordinate frame and is thus oriented downwards. We can clearly see that the effective rotations relative to the objects' orientation perfectly align, demonstrating that the learned embeddings correctly encode 3D poses and rotations. For the 3D Blocks and Reacher, we train a separate decoder for $100$ epochs after freezing our model in order to decode $z$ into pixel space. Figures~\ref{fig:cubes_decoder} and \ref{fig:reacher_decoder} show our model implicitly learns a reasonable group action $\rho_\gS$ in input space which corresponds the group action in latent space $\rho_\gZ$.

\vspace{-0.25\baselineskip}
\subsection{Generalization from limited actions}
\label{exp:generalization}
In these experiments, we train on a subset of actions and evaluate on datasets generated with the full action space. These experiments aim to verify that our model, even with components not constrained to be equivariant, can learn a good equivariant representation which can generalize to actions that were not seen during training. We perform experiments on the 2D Shapes, 3D Blocks, and Reacher domains. For 2D Shapes, the training data contains only `up' actions and for 3D Blocks, we omit the left action in training. For Reacher, we restrict the actuation force of the second joint to be positive, meaning that the second arm rotates in only one direction.

Tables~\ref{tab:blocks_skew} and \ref{tab:reacher_skew} show results for 2D Shapes, 3D Blocks, and Reacher. We see that our method can successfully generalize over unseen actions compared to both the non-equivariant and fully equivariant baselines. The non-equivariant baseline in particular performs poorly on all domains, achieving only 2.8\% on Hits@1 and 5.3\% on MRR for 2D Shapes. The fully equivariant model performs worse than our method for 3D Blocks and achieves a similar performance on Reacher. As the fully equivariant model performs well when trained on all actions but does not perform as well in these generalization experiments, these results lend support to our hypothesis that the inaccurate pixel-level equivariance bias limits its extrapolation abilities to out-of-distribution samples. In these limited actions experiments, the fully equivariant model cannot extrapolate correctly.

Figure~\ref{fig:2d_shapes_viz} in the Appendix shows embeddings for all states in the evaluation dataset for our model and the non-equivariant model trained on only the up action. Our model 
shows a clear $5 \times 5$ grid, while the non-equivariant model learns a degenerate solution (possibly encoding only the row index).

\vspace{-0.25\baselineskip}
\section{Conclusion and Future Work}

This work demonstrates that an equivariant world model can be paired with a symmetric embedding network, which itself is not equivariant by construction, to learn a model that is approximately equivariant. This makes it possible to use equivariant neural networks in domains where the symmetry is known, but transformation properties of the input data cannot be described explicitly. We consider a variety of domains and equivariant neural network architectures, for which we demonstrate generalization to actions outside the training distribution.
Future work will include tasks besides world models and using symmetric embeddings to develop disentangled and more interpretable features in domains with known but difficult to isolate symmetry. Other possible avenues include standardizing the proposed meta-architecture for wider accessibility and using other non-convolutional SEN architectures for different data types such as point clouds and graphs.

\section*{Acknowledgements}

This work was supported by NSF Grants \#2107256 and \#2134178. R. Walters is supported by The Roux Institute and the Harold Alfond Foundation. We thank Lawson Wong and the anonymous reviewers for the helpful discussion and feedback on the paper. This work was completed in part using the Discovery cluster, supported by Northeastern University’s Research Computing team.

\nocite{tange_ole_2018_1146014}
\bibliography{references}
\bibliographystyle{icml2022}

\newpage
\appendix
\onecolumn

\section{Outline}
Our appendix is organized as follows.
First, in Section \ref{app:setup}, we provide an additional details on the problem setup.
Additional experiment results are presented in Section \ref{app:experiments}, followed by the details of environments and network architectures in Sections \ref{app:environ} and \ref{app:model}.
We further explain the notation and definition in Section \ref{app:reps}.  The proof of Proposition \ref{prop:gen-limited-actions} is in Section \ref{app:prop-proof}.

\section{Setup: Equivariant World Models}
\label{app:setup}

\label{app:mdp_symmetry}
In this section, we provide a technical background for building equivariant world models, which we use in learning symmetric representations.

We model our interactive environments as Markov decision processes. A (deterministic) Markov decision process (MDP) is a $5$-tuple $\mathcal{M}=\langle \mathcal{S}, \mathcal{A}, T, R, \gamma \rangle$, with state space $\mathcal{S}$, action space $\mathcal{A}$, (deterministic) transition function $T: \mathcal{S} \times \mathcal{A} \to \mathcal{S}$, reward function $R: \mathcal{S} \times \mathcal{A} \to \mathbb{R}$, and discount factor $\gamma \in [0, 1]$.

Symmetry can appear in MDPs naturally \citep{zinkevich2001symmetry,narayanamurthy2008hardness,ravindran2004algebraic,van2020mdp}, which we can exploit using equivariant networks.
For example, \citet{van2020mdp} study geometric transformations, such as reflections and rotations.
\citet{ravindran2004algebraic} study group symmetry in MDPs as a special case of MDP homomorphisms.

\paragraph{Symmetry in MDPs.}

Symmetry in MDPs is defined by the automorphism group $\mathrm{Aut}(\mathcal{M})$ of an MDP, where an automorphism $g \in \mathrm{Aut}(\mathcal{M})$ is an MDP homomorphism $h: \mathcal{M} \to \mathcal{M}$ that maps $\mathcal{M}$ to itself and thus preserves its structure.
\citet{zinkevich2001symmetry} show the invariance of value function for an MDP with symmetry.
\citet{narayanamurthy2008hardness} prove that finding exact symmetry in MDPs is graph isomorphism complete.

%
\citet{ravindran2004algebraic} provide a comprehensive overview of using MDP homomorphisms for state abstraction and study symmetry in MDPs as a special case.
A more recent work by \citet{van2020mdp} builds upon the notion of MDP homomorhpism induced by group symmetry and uses it in an inverse way. They assume knowledge of MDP homomorphism induced by symmetry group is known and exploit it.
Different from us, their focus is on policy learning, which needs to preserve both transition and reward structure and thus has \textit{optimal value equivalence} \citep{ravindran2004algebraic}.

%
More formally, an \textit{MDP homomorphism} $h: \mathcal{M} \to \overline{\mathcal{M}}$ is a mapping from one MDP $\mathcal{M}=\langle \mathcal{S}, \mathcal{A}, T, R, \gamma \rangle$ to another $\overline{\mathcal{M}}=\langle \overline{\mathcal{S}}, \overline{\mathcal{A}}, \overline{T}, \overline{R}, \gamma \rangle$ which needs to preserve the transition and reward structure~\citep{ravindran2004algebraic}.  
The mapping 
$h$ consists of a tuple of surjective maps $h=\langle \phi, \{ \alpha_s \mid s \in \mathcal{S} \} \rangle$, where $\phi: \mathcal{S} \to \overline{\mathcal{S}}$ is the state mapping and $\alpha_s: \mathcal{A} \to \overline{\mathcal{A}}$ is the \textit{state-dependent} action mapping.
The mappings are constructed to satisfy the following conditions:
(1) the transition function is preserved 
$
\overline{T}\left(\phi\left(s^{\prime}\right) \mid \phi(s), \alpha_{s}(a)\right) = \sum_{s^{\prime \prime} \in \phi^{-1}\left(\phi(s^{\prime})\right)} T\left(s^{\prime \prime} \mid s, a\right)
$,
(2) and the reward function is also preserved 
$
\overline{R}\left(\phi(s), \alpha_{s}(a)\right) = R(s, a),
$
for all $s, s^{\prime} \in \mathcal{S}$ and for all $a \in \mathcal{A}$.

%
An MDP isomorphism from an MDP $\mathcal{M}$ to itself is call an \textit{automorphism} of $\mathcal{M}$.
The collection of all automorphisms of $\mathcal{M}$ along with the composition of homomorphisms is the \textit{automorphism group} of $\mathcal{M}$, denoted $\mathrm{Aut}(\mathcal{M})$.


We specifically care about a subgroup of $G \subseteq \mathrm{Aut}(\mathcal{M})$ which is usually easily identifiable from environments a priori and thus we can design appropriate equivariant network architectures to respect it, such as $C_4$ rotation symmetry of objects.
Additionally, while MDP homomorphisms pose constraints to the transition and reward function, we only care about the transition function, especially the \textit{deterministic} case $T: \mathcal{S} \times \mathcal{A} \to \mathcal{S}$.

\paragraph{Equivariant transition.}

By definition, when an MDP $\mathcal{M}$ has symmetry, any state-action pair $(s,a)$ and its transformed counterpart $(\rho_{\mathcal{S}}(g) \cdot s, \rho_{\mathcal{A}}(g) \cdot a) $ are mapped to the same abstract state-action pair by $h \in \mathrm{Aut}(\mathcal{M})$: $(\phi(s), \alpha_{s}(a)) = (\phi(gs), \alpha_{gs}(ga))$, for all $s \in \mathcal{S}, a \in \mathcal{A}, g \in G$.
Therefore, the transition function $T: \mathcal{S} \times \mathcal{A} \to \mathcal{S}$ should be $G$-equivariant: 
\begin{equation}
    T(\rho_\mathcal{S}(g) \cdot s, \rho_\mathcal{A}(g) \cdot a) = \rho_\mathcal{S}(g) \cdot T(s, a),
\end{equation} 
for all $s \in \mathcal{S}, a \in \mathcal{A}, g \in G$.

\paragraph{State-dependent action transformation.}
\label{para:action-transformation}

Note that
the group operation acting on action space $\mathcal{A}$ \textit{depends on state}, since $G$ actually acts on the \textit{product space} $\mathcal{S} \times \mathcal{A}$: $(g, (s,a)) \mapsto \rho_{\mathcal{S}\times\mathcal{A}}(g) \cdot (s,a)$.
However, in most cases, including all of our environments, the action transformation $\rho_\mathcal{A}$ does not depend on state.
As a bibliographical note, the formulation in \citet{van2020mdp} also has a joint group action on state and action space, which is denoted as state transformation $L_g: \mathcal{S} \to \mathcal{S}$ and \textit{state-dependent} action transformation $K_g^s: \mathcal{A} \to \mathcal{A}$.
Table 1 in \citet{van2020mdp} outlines state and action transformations for their environments, and all of action transformations are not state-dependent.

Similarly in our case, geometric transformations are usually acting globally on the environments $\mathcal{S} \times \mathcal{A}$, thus states and actions are transformed accordingly.
We use the factorized form and omit the state-dependency $\rho_\mathcal{A}(g;s)$ of action transformation $\rho_\mathcal{A}(g)$, since the action transformations do not depend on states $\rho_\mathcal{A}(g;s) = \rho_\mathcal{A}(g)$ for all $g \in G$, $s \in \mathcal{S}$.

%

\paragraph{Learning transition with equivariant networks.}

In this paper we are mainly interested in learning transition functions which are equivariant under symmetry transformations and can be high-dimensional.

We apply the idea of learning equivariant transition models in the latent space $\mathcal{Z}$, where $\mathcal{Z}$ is the space of symmetric representations, on various environments with different symmetry groups $G$.
We assume we do not explicitly know $\rho_\gS$ since $\gS$ is high-dimensional.
We factorize the group representation on state and action $\mathcal{S} \times \mathcal{A}$ as latent state transformation $\rho_{\mathcal{Z}}(g) \cdot E(s)$ and $\rho_\mathcal{A}(g;s) \cdot a$.
In the deterministic case, the transition model can be modeled by $G$-equivariant networks in latent state $\gZ$ and action space $\mathcal{A}$:
%
\begin{equation}
    \rho_\mathcal{Z}(g)\cdot T(E(s), a) = T(\rho_\mathcal{Z}(g)\cdot E(s), \rho_\mathcal{A}(g)\cdot a),
\end{equation}
for all $g \in G$, $s, s' \in \mathcal{S}$ and $a \in \mathcal{A}$.

\section{Experiment Results}
\label{app:experiments}

\subsection{Model performance comparison}
Tables~\ref{tab:blocks_rush_hour},\ref{tab:reacher} show model comparison results for 3D Blocks, Rush Hour, and Reacher. All models generally perform well on the accuracy metrics. The fully-equivariant model performs well on the equivariance metrics and our model performs better than the non-equivariant model.

\begin{table*}[!h]
\centering
\begin{tabular}{llrrrr}
\toprule
& Model & \multicolumn{1}{c}{\begin{tabular}{@{}c@{}} Hits@1 \\ (10 step, \%)\end{tabular}} & \multicolumn{1}{c}{\begin{tabular}{@{}c@{}}MRR \\ (10 step, \%)\end{tabular}} & \multicolumn{1}{c}{EE($S$)} & \multicolumn{1}{c}{\begin{tabular}{@{}c@{}} DIE \\ (10 step, $\times 10^{-2}$)\end{tabular}} \\ 
\midrule
\multirow{3}{*}{3D Blocks} & None & $94.3{\color{gray}\scriptstyle \pm 9.0}$ & $99.0{\color{gray}\scriptstyle \pm 1.5}$ & \multicolumn{1}{r}{$0.89{\color{gray}\scriptstyle \pm 0.3}$} &  \multicolumn{1}{r}{$3.85{\color{gray}\scriptstyle \pm 2.0}$} \\
& Full & $99.8{\color{gray}\scriptstyle \pm 0.3}$ & $99.9{\color{gray}\scriptstyle \pm 0.2}$ & \multicolumn{1}{r}{$0.82{\color{gray}\scriptstyle \pm 0.5}$} & \multicolumn{1}{r}{$0.55{\color{gray}\scriptstyle \pm 0.5}$} \\
& Ours & $99.9{\color{gray}\scriptstyle \pm 0.0}$ & $100{\color{gray}\scriptstyle \pm 0.0}$ & \multicolumn{1}{r}{$0.86{\color{gray}\scriptstyle \pm 0.4}$} & \multicolumn{1}{r}{$0.32{\color{gray}\scriptstyle \pm 0.2}$} \\ 
\midrule
\multirow{3}{*}{Rush Hour} & None & 98.2${\color{gray}\scriptstyle \pm 1.3}$ & $99.1{\color{gray}\scriptstyle \pm 0.7}$ &  $0.37{\color{gray}\scriptstyle \pm 0.07}$ &  $26.6{\color{gray}\scriptstyle \pm 7.13}$ \\
& Full & $87.9{\color{gray}\scriptstyle \pm 3.1}$ & $93.6{\color{gray}\scriptstyle \pm 1.7}$ & $0.00{\color{gray}\scriptstyle \pm 0.00}$ & $0.05{\color{gray}\scriptstyle \pm 0.06}$ \\
& Ours & $93.6{\color{gray}\scriptstyle \pm 3.7}$ & $96.1{\color{gray}\scriptstyle \pm 2.0}$ & $0.26{\color{gray}\scriptstyle \pm 0.09}$ & $10.0{\color{gray}\scriptstyle \pm 3.45}$ \\
\bottomrule
\end{tabular}
\caption{Model performance on 3D Blocks and Rush Hour.}
\label{tab:blocks_rush_hour}
\end{table*} 

\begin{table*}[!h]
\centering
\begin{tabular}{llrrrr}
\toprule
& Model & \begin{tabular}{@{}c@{}} H@10 \\ (1 step, \%)\end{tabular} & \begin{tabular}{@{}c@{}}MRR \\ (1 step, \%)\end{tabular} & EE($S$) & \begin{tabular}{@{}c@{}} DIE \\ (1 step, $\times 10^{-1}$)\end{tabular} \\
\midrule
\multirow{3}{*}{Reacher} & None & $100{\color{gray}\scriptstyle \pm 0.0}$ & $88.3{\color{gray}\scriptstyle \pm 3.3}$ & $1.26{\color{gray}\scriptstyle \pm 0.1}$ & $5.61{\color{gray}\scriptstyle \pm 1.8}$ \\
& Full & $100{\color{gray}\scriptstyle \pm 0.0}$ & $95.5{\color{gray}\scriptstyle \pm 1.9}$ & $1.19{\color{gray}\scriptstyle \pm 0.0}$ & $3.91{\color{gray}\scriptstyle \pm 0.9}$ \\
& Ours & $100{\color{gray}\scriptstyle \pm 0.0}$ & $94.1{\color{gray}\scriptstyle \pm 2.8}$ & $1.29{\color{gray}\scriptstyle \pm 0.0}$ & $5.23{\color{gray}\scriptstyle \pm 0.8}$ \\
\bottomrule
\end{tabular}
\caption{Model performance on Reacher.}
\label{tab:reacher}
\end{table*}

\subsection{Latent visualizations}
\label{app:latent-repr}

\begin{figure}[H]
    \centering
     \begin{subfigure}[b]{0.215\textwidth}
        \centering
        \includegraphics[width=\textwidth]{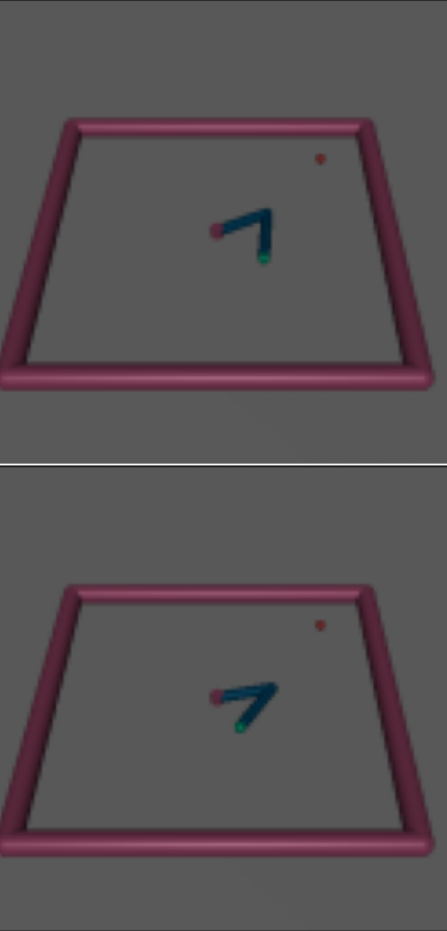}
        \caption{$s, s'$}
     \end{subfigure}
     \hfill  
    \begin{subfigure}[b]{0.7\textwidth}
        \centering
        \includegraphics[width=\textwidth,trim=100 70 85 65,clip]{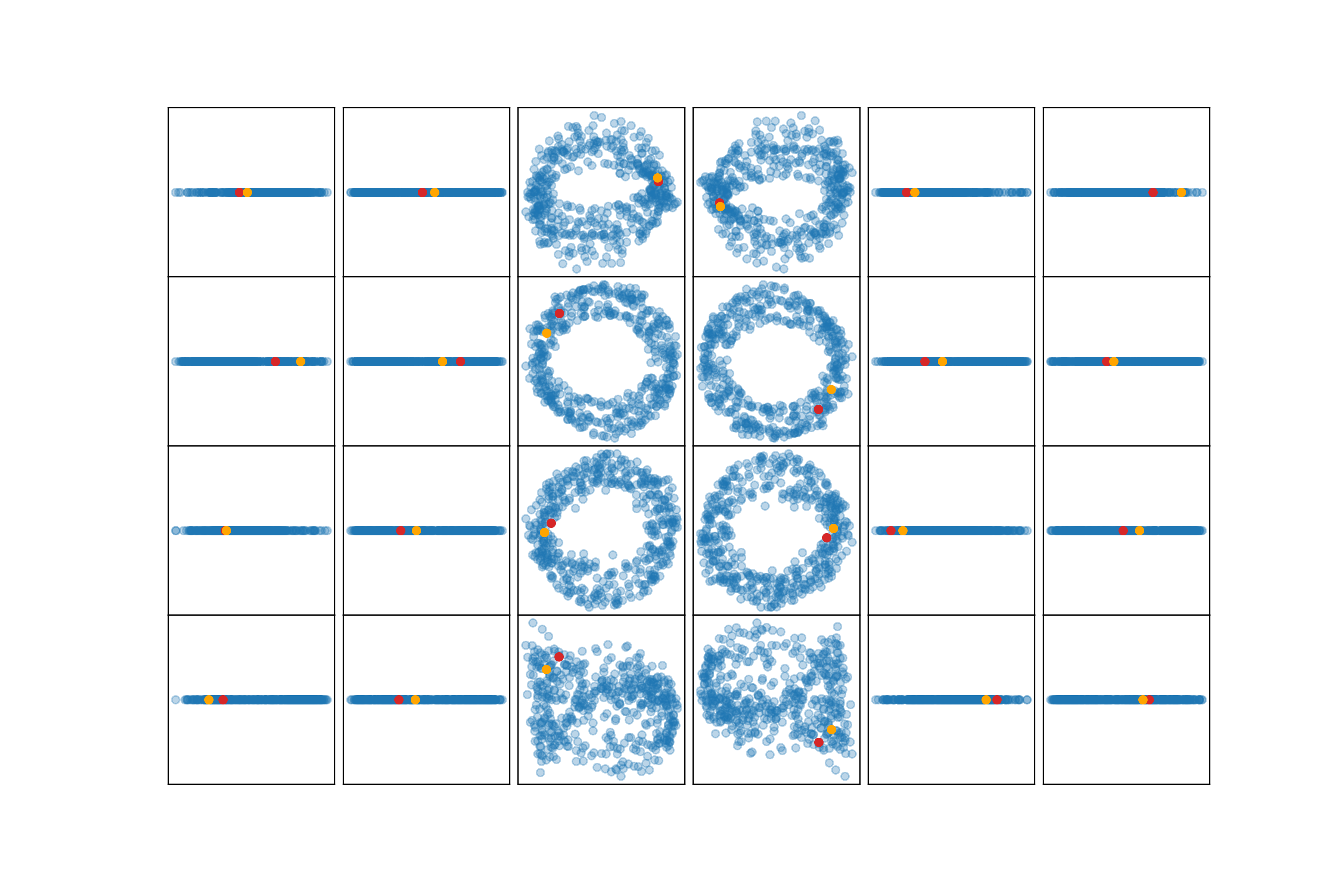}
        \caption{$z \in \gZ$.  }
     \end{subfigure}
    \caption{Learned symmetric embeddings for Reacher: pixel observation $s$ (left top) and next observation $s'$ (left bottom), latent representation $z$ of the evaluation set (right). The representation type of $z$ is $\rho_{D_4,\mathrm{reg}}$ which we factor into irreducible representations before visualizing (see \cite{hall2003lie}). All encoded samples in the evaluation set are shown and the encoded current observation is colored red and the encoded next observation is colored orange. There is a clear circular pattern that match joint rotations.}
    \label{fig:reacher_z}
\end{figure}

\begin{figure}[H]
    \centering
     \begin{subfigure}[b]{0.3\textwidth}
        \centering
        \includegraphics[width=\textwidth]{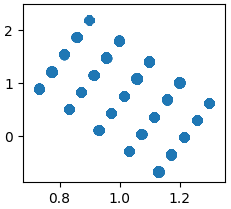}
        \caption{}
     \end{subfigure}
    \begin{subfigure}[b]{0.3\textwidth}
        \centering
        \includegraphics[width=\textwidth]{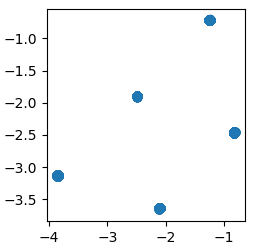}
        \caption{}
     \end{subfigure}
    \caption{2D Shapes: learned embeddings for all states in the evaluation set when trained on only the up action. Our model (left) is able to generalize well and learns the correct underlying $5 \times 5$ grid. The non-equivariant model (right) learns a degenerate solution}
    \label{fig:2d_shapes_viz}
\end{figure}

The learned latent embedding $z$ for all states in the evaluation set for Reacher is shown in Figure~\ref{fig:reacher_z}. The embeddings are factored into irreducible representations. The 2-dimensional representations show a circular pattern, mimicking the rotation of joints.  Figure~\ref{fig:2d_shapes_viz} shows learned embeddings for all states in the evaluation dataset of 2D Shapes for the generalization experiments, where the models were trained on only the up action.  Our model (left) correctly encodes the underlying $5\times5$ grid while the non-equivariant model (right) encodes a degenerate solution.

\subsection{Decoder reconstructions}
\begin{figure}[H]
    \centering
    \includegraphics[width=0.9\textwidth]{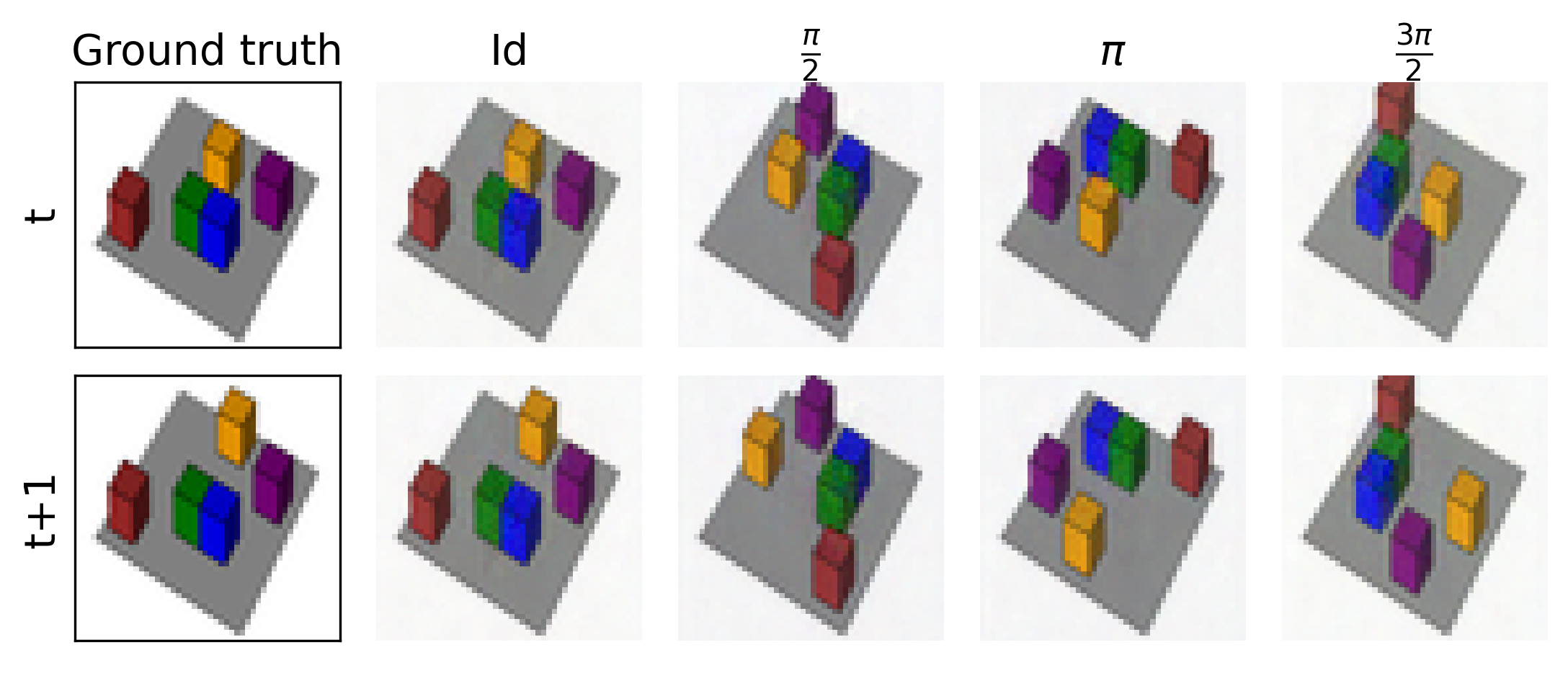}
    \caption{3D Blocks: Pixel-level reconstructions of the $G$-transformed latent representations. The $C_4$ symmetry acts correctly in the input space $\gS$}
    \label{fig:cubes_decoder}
\end{figure}

\begin{figure}[H]
    \centering
    \includegraphics[width=\textwidth]{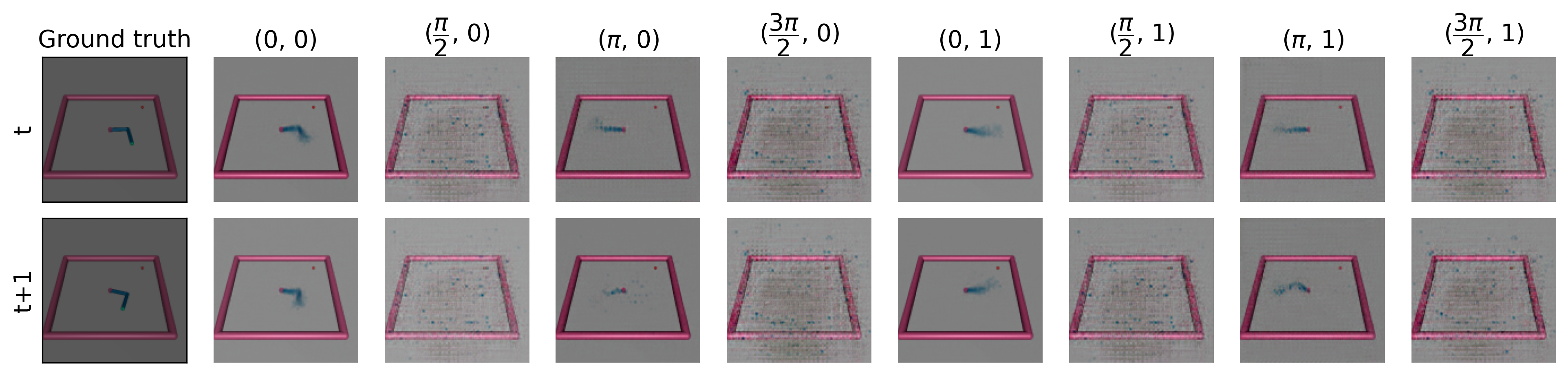}
    \caption{Reacher: Pixel-level reconstructions of the $G$-transformed latent representations. We denote the representations in $D_4$ as a tuple of $\pi/2$ rotations and a reflection (0 or 1). While many reconstructions are blurry, rotations of $\pi$ and reflection seem to act correctly in the input space $\gS$.}
    \label{fig:reacher_decoder}
\end{figure}

We freeze our model and train a separate decoder for $100$ epochs (until convergence) for 3D Blocks and Reacher environments. Having a trained decoder allows us to qualitatively assess the learned latent representations and further allows us to transform the representations with the group actions $\rho_\gZ$ and visualize their pixel-level outputs. These results show that our model correctly infers and transforms group actions in the input space $\rho_\gS$ and translates them to group actions in latent space $\rho_\gZ$.

\section{Datasets and Environments}
\label{app:environ}

\begin{figure}[!t]
    \centering
    \includegraphics[width=0.5\columnwidth]{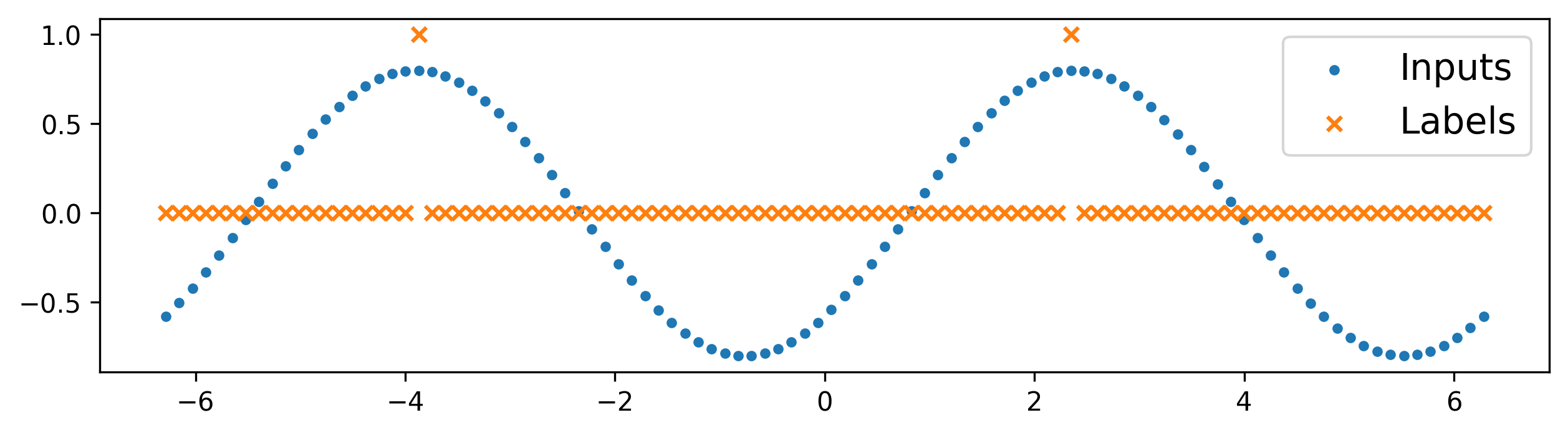}
    \caption{Sequence labeling of local maxima in sine waves. Each input and label sample consist of $100$ points.}
    \label{fig:sine1D_data}
\end{figure}

\paragraph{Sequence labeling}
A sample time series containing $100$ time points and its corresponding labels is shown in Figure~\ref{fig:sine1D_data}. 

A random policy was used to create training and evaluation datasets of $(s, a, s')$ tuples. For all environments, we have either a combinatorially large state space (with objects) or continuous states and thus overlap between training and evaluation datasets is highly unlikely. 

\paragraph{2D Shapes \& 3D Blocks}
There are five objects are arranged in a $5 \times 5$ grid and each object can occupy a single cell. Actions are the 4 cardinal directions for each object and an action moves one object at a time, unless it is blocked by the boundaries or by another object. Observations are $50 \times 50 \times 3$ RGB images for both 2D Shapes and 3D Blocks, with pixel values normalized to $[0,1]$. The observations in 2D shapes are top down views of the grid and each object has a different color-shape combination. For 3D Blocks, the observations are rendered isometrically with a skewed perspective and each block has a $z$-height, introducing partial occlusions to the image.

\paragraph{Rush Hour}
We create a variant of 2D Shapes called Rush Hour. Each object has an orientation and the action is relative to the object's orientation: $\{ \text{forward}$, $\text{backward}$, $ \text{left}$, $\text{right}\}$.  This increases the importance of rotational orientation in the environment increasing the significance of symmetric embeddings.
We fix the color of all objects and use randomized orientation $\text{north}$, $\text{west}$, $\text{east}$, $\text{south}\}$ for each object at each episode.

\paragraph{Reacher}
This environment makes a small modification to the original MuJoCo environment \texttt{Reacher-v2}. As we do not consider rewards, we fix the goal position to the position $[0.2,0.2]$ so that features related to the goal are ignored. Instead of using the $11$-dimensional state, we use pixel observations as images and preprocess them by cropping slightly and downsampling the original $500 \times 500 \times 3$ RGB image to $128 \times 128 \times 3$. The previous and current frames are then stacked as an observation to encode velocities. The default camera position gives a slightly skewed perspective, see Table~\ref{tab:exp_summary}.

\paragraph{3D Teapot}
The 3D teapot environment contains images of the Utah teapot model rendered into the $64 \times 64$ grayscale images.  The teapot varies in pose which can be described by a coordinate frame $z \in \mathrm{SO}(3)$. Actions may be any element $a \in \mathrm{SO}(3)$.  


\begin{figure}[b]
     \begin{subfigure}[b]{0.4\textwidth}
        \begin{subfigure}[b]{0.49\textwidth}
         \centering
         \includegraphics[width=\textwidth]{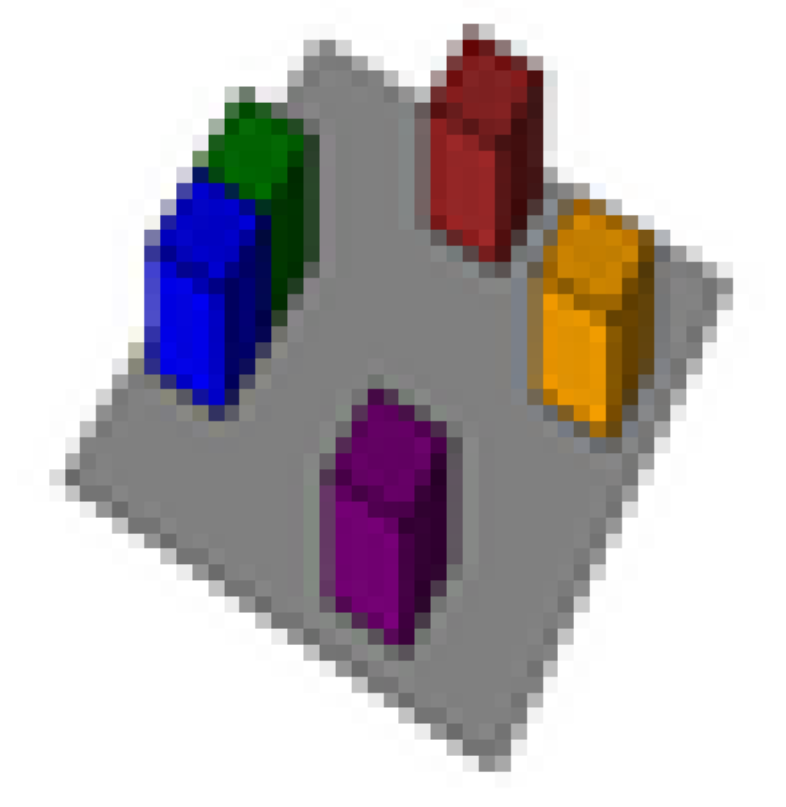}
        \end{subfigure}
        \begin{subfigure}[b]{0.49\textwidth}
         \centering
         \includegraphics[width=\textwidth]{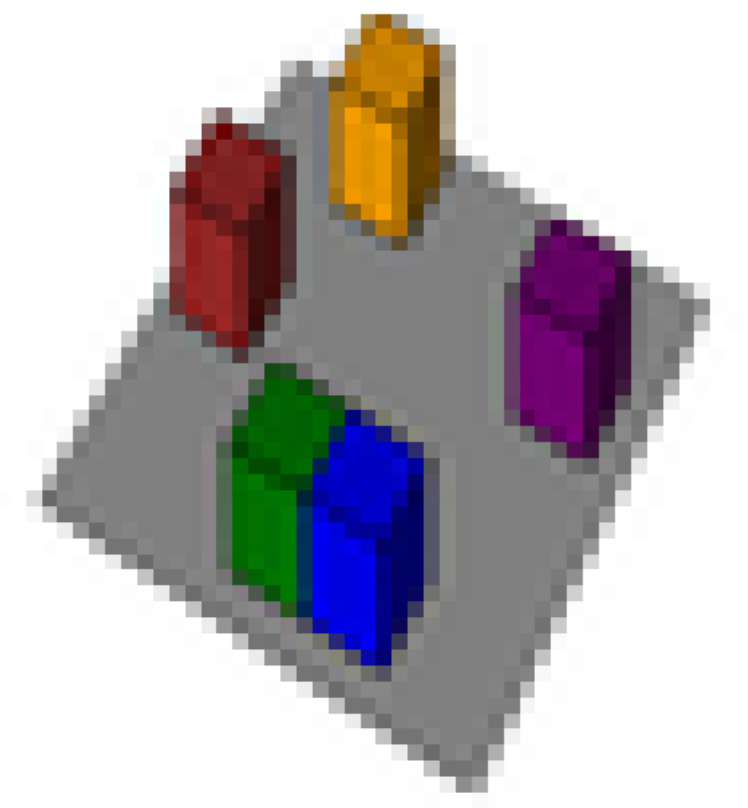}
        \end{subfigure}
        \caption{3D Blocks}
     \end{subfigure}
     \hfill
        \begin{subfigure}[b]{0.4\textwidth}
        \begin{subfigure}[b]{0.49\textwidth}
         \centering
         \includegraphics[width=\textwidth]{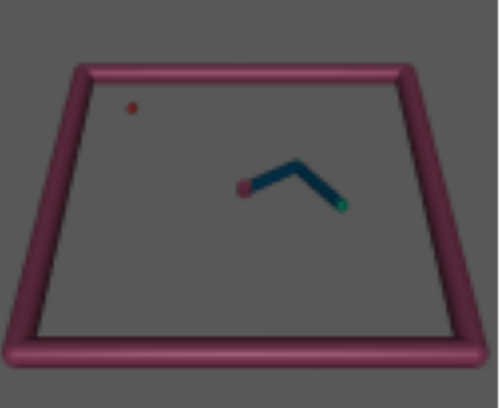}
        \end{subfigure}
        \begin{subfigure}[b]{0.49\textwidth}
         \centering
         \includegraphics[width=\textwidth,trim={0 12 0 15},clip]{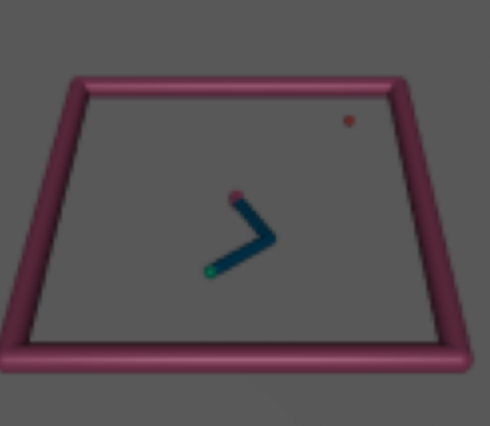}
        \end{subfigure}
        \caption{Reacher}
     \end{subfigure}
    \caption{Original observations and their $G$-transformed versions}
    \label{fig:rot_environments}
\end{figure}

\section{Model architectures}
\label{app:model}

\paragraph{Symmetric embedding network $S$} For all models and environments except for 3D Teapot, we use CNNs with BatchNorm \cite{pmlr-v37-ioffe15} and ReLU activations between each convolutional layer. For 3D Teapot, the symmetric embedding network maps directly to the latent $z$ space so we use $4$ convolutional layers followed by $3$ fully connected layers. The output is a $3 \times 3$ rotation matrix. The number of layers for each environment is given in Table~\ref{tab:exp_summary}. For the non-equivariant symmetric embedding networks, we use $32$ convolutional filters for every layer and use $8$ filters for Reacher and $16$ filters for all other environments. 

\paragraph{Encoder $E$} The object-oriented environments use 3-layer MLPs with $512$ hidden units for the non-equivariant networks and $256$ for the equivariant counterparts. There is a ReLU activation after the first and second layers and a LayerNorm \citep{ba2016layer} after the second layer. For Reacher, we use 3 convolutional layers followed by 3 fully connected layers. The 3D Teapot does not have an explicit encoder, i.e. it is the identity function. The output of the non-equivariant encoder is a 2-dimensional vector for 2D Shapes, 3D Blocks, and Rush Hour and a 4-dimensional vector for Reacher. The output of the equivariant encoders for each environment is listed in Table~\ref{tab:exp_summary}.


\paragraph{Transition $T$} The object-oriented environments use message-passing transition models where the edge and node networks have the same structure as the encoder (3-layer MLPs). For Reacher and 3D Teapot, the transition model $T$ is a MLP with $512$ hidden units for the non-equivariant version and $256$ for the equivariant version. Actions are concatenated to the latent $z$ and are input into the transition models which then outputs a $z'$ of the same dimension as the input $z$. We use one-hot encoding for discrete actions.

\paragraph{Gram-Schmidt embedding for Teapot transition model}
In the case of the teapot domain, the transition model is constrained to output an element of $\mathrm{SO}(3)$ representing a positively-oriented orthonormal frame.  This is achieved by having the network output two vector $u ,v \in \mathbb{R}^3$ and then performing Gram-Schmidt orthogonalization.
Only two vectors are necessary since orthogonality and orientation determine the third, 
 after producing two orthonormal vectors $u',v'$, the third vector $w'$ is uniquely determined by the property that it completes a positively-oriented orthonormal frame and can be computed by cross product.  In summary,
\begin{align*}
    u' = u / \|u\|,  &\qquad\qquad
    v' = \frac{v - (u' \cdot  v) u'}{\| v - (u' \cdot  v) u'\| }, \\
    w' = u' \times v', &\qquad\qquad
    y = [u'\ v'\ w'].
\end{align*}

\section{Training details}
\label{app:training}

For training, we use $1000$ episodes of length $100$ as training data for the grid world environments (2D shapes, 3D blocks, Rush Hour), $2000$ episodes of length $10$ for Reacher, and $100{,}000$ episodes of length $1$ for the 3D teapot. For Reacher, the starting state is restricted to a subset of the whole state space, so we perform warm starts with $50$ random actions in order to generate more diverse data. The evaluation datasets are generated with different seeds from the training data to ensure that transitions are different. In the generalization experiments for 2D Shapes and 3D blocks, we set the number of episodes in the training data to $100{,}000$ with length $1$ to avoid any distribution shifts in the data (e.g. performing up continuously will produce many transitions where all blocks are blocked by the boundaries).

For the object-oriented environments, we follow the hyperparameters used in \citep{kipf20}: a learning rate of $5 \times 10^{-4}$, batch size of $1024$, $100$ epochs, and the hinge margin $\gamma = 1$. We find that these hyperparameters work well for all other environments, except that Reacher uses a batch size of $256$ and mixed precision training was used for both non-equivariant, fully-equivariant, and our method, in order to keep the batch size relatively high for stable contrastive learning. Most experiments were run on a single Nvidia RTX 2080Ti except for 3D Blocks which used a single Nvidia P100 12GB.

\section{Group Representations}
\label{app:reps}

We explain the notation and definitions of the different representations of the groups considered in the paper and displayed in Table \ref{tab:exp_summary}.

The $\rho_\mathrm{std}$ representation of $C_4$ or $D_4$ on $\mathbb{R}^2$ is by 2-by-2 rotation and reflection matrices. The $\rho_\mathrm{std}$ representation of $S_5$ permutes the standard basis of $\mathbb{R}^5$.  The regular representation $\rho_{\mathrm{reg}}$ of $G$ permutes the basis element of $\mathbb{R}^{|G|}$ according to the multiplication table of $G$.  The trivial representation of $G$ fixes $\mathbb{R}$ as $\rho_{\mathrm{triv}}(g) \mvdot x = x$. For $D_4$, $\rho_{\mathrm{flip}}(g) = \pm 1$ is a representation on $\mathbb{R}$ depending only on if $g$ contains a reflection. Given representations $(\rho_1,\mathbb{R}^{n_1})$ and $(\rho_2,\mathbb{R}^{n_2})$ of $G_1$ and $G_2$, $(\rho_1 \boxtimes \rho_2)(g_1,g_2)(v \otimes w) = g_1 v \otimes g_2 w$ gives a representation on $G_1 \times G_2$ on $\mathbb{R}^{n_1} \otimes \mathbb{R}^{n_2}$.

\section{Proof of Proposition \ref{prop:gen-limited-actions}}
\label{app:prop-proof}

\begin{proposition*}[\ref{prop:gen-limited-actions}]
Let $\gA' \subset \gA$.  Assume $\rho_\gA(G) \mvdot \gA' = \gA$, i.e. every MDP action is a $G$-transformed version of one in $\gA'$.  Consider $\gD'$ sampled from $\gS \times \gA' \times \gS$.  
Denote $\gD = G \cdot \gD' \subset \gS \times \gA \times \gS$ the set of all $G$-transforms of all of samples in $\gD'$. Assume a $G$-invariant norm.
Denote the full model $T_\gS(s,a) = T(E(S(s)),a)$.   Assume model error is bounded $\| T_\gS(s,a) - z' \| \leq \epsilon_1$ where $z' = E(S(s'))$ and 
equivariance errors are bounded $\| T_\gS(\rho_\gS(g)s,\rho_\gA(g)a) - \rho_\gZ(g) T_\gS(s,a)  \| \leq \epsilon_2$ and $\| E(S(\rho_\gS(g)s') - \rho_\gZ(g) z' \| \leq \epsilon_3$ for all $g \in G$ and all $(s,a,s') \in \gD'$.  Then model error on the full action space is also bounded $\| T_\gS(s,a) - z' \| \leq \epsilon_1 + \epsilon_2 + \epsilon_3$ for all $(s,a,s') \in \gD$. 
\end{proposition*}

\begin{proof}
Let $(s_1,a_1,s_1') \in \gD$.  Then there exists $(s,a,s') \in \gD'$ such that $(\rho_\gS(g) s , \rho_\gA(g) a, \rho_\gS(g) s' ) = (s_1, a_1, s_1')$.  Let $z' = E(S(s'))$ and $z_1' = E(S(s_1'))$.    By triangle inequality
\begin{align*}
    \|  T_\gS(s,a) - z' \| &= 
    \|  T_\gS(\rho_\gS(g^{-1})s_1,\rho_\gA(g^{-1})a_1) - z' \| \\
    &\leq  \|  T_\gS(\rho_\gS(g^{-1})s_1,\rho_\gA(g^{-1})a_1) - \rho_\gZ(g^{-1}) T_\gS(s_1,a_1)\| + \|\rho_\gZ(g^{-1}) T_\gS(s_1,a_1) - \rho_\gZ(g^{-1})z_1' \| \\
    &\quad + \| \rho_\gZ(g^{-1}) z_1' - z' \|
\end{align*}
By the bound on equivariance error, the first term is bounded by $\epsilon_2$. The third term is bounded 
\[
\| \rho_\gZ(g^{-1}) z_1' - z' \| = \| \rho_\gZ(g^{-1})E(S(s_1')) - E(S(\rho_\gS(g^{-1}) s_1')) \| \leq \epsilon_3.
\]
By invariance of the norm, the middle term
\[
\|\rho_\gZ(g^{-1}) T_\gS(s_1,a_1) - \rho_\gZ(g^{-1})z_1' \|  = \|T_\gS(s_1,a_1) - z_1' \| \leq \epsilon_1 
\]
Combining the bounds yields the result.
\end{proof}

If $S$ learns to be equivariant, then the composite models $T(E(S(\cdot),\cdot))$ and $E(S(\cdot))$ will be equivariant, thus minimizing $\epsilon_2$ and $\epsilon_3$.  Minimizing $\epsilon_1$ is part of the objective. Thus low error on unseen MDP actions $\gA$ is feasible. 

The assumption the norm $\| \cdot \|$ is $G$-invariant is valid for, e.g., rotations, reflections, and permutations.  If the norm is not invariant, then the proof goes through with the modified loss bound $\epsilon_1 \| g^{-1} \| + \epsilon_2 + \epsilon_3$ where $\| g\|$ denotes the operator norm of the element $g \in G$ which relates the sample to one in $\gD'$.

\end{document}

%% file: math_commands.tex
\usepackage{amsmath,amsfonts,bm}

\def\eqref#1{equation~\ref{#1}}

\def\1{\bm{1}}

\DeclareMathAlphabet{\mathsfit}{\encodingdefault}{\sfdefault}{m}{sl}
\SetMathAlphabet{\mathsfit}{bold}{\encodingdefault}{\sfdefault}{bx}{n}

\def\gA{{\mathcal{A}}}

\def\gD{{\mathcal{D}}}

\def\gS{{\mathcal{S}}}

\def\gY{{\mathcal{Y}}}
\def\gZ{{\mathcal{Z}}}

